\documentclass{article}
\usepackage[numbers]{natbib}
\usepackage{PRIMEarxiv}
\usepackage[T1]{fontenc}
\usepackage[utf8]{inputenc}
\usepackage[american]{babel}
\usepackage{csquotes}
\usepackage{microtype}
\usepackage[defblank]{paralist}
\setdefaultenum{(1)}{(a)}{(i)}{A.}
\usepackage[inline]{enumitem}
\usepackage{url}
\usepackage{flushend}
\usepackage{etoolbox}
\usepackage{fix-cm}
\usepackage{textpos}
\usepackage{mfirstuc}
\usepackage{titlecaps}
\usepackage[datesep=.,style=ddmmyyyy]{datetime2}
\usepackage[noend]{algpseudocode}

\usepackage{algorithm}
\usepackage{algpseudocode}
\usepackage{pgf}
\newcommand{\newPerc}[1]{#1\%}

\newcommand{\perc}[1]
{\pgfmathparse{#1*100}\pgfmathprintnumber{\pgfmathresult}\%}
\usepackage{tcolorbox}

\newtcolorbox{Summary}{
    sharpish corners, 
    boxrule = 0pt, 
    toprule = 3.5pt, 
    toptitle = 1mm, 
    enhanced,
    fuzzy shadow = {0pt}{-2pt}{-0.5pt}{0.5pt}{black!35}, 
    colback = white, 
    colframe = gray, 
    coltitle=black, 
    fonttitle=\bfseries 
}
\usepackage[framemethod=tikz]{mdframed}

\mdfdefinestyle{mpdframe}{
    frametitlebackgroundcolor   =black!15,
    frametitlerule              =true,
    roundcorner                 =5pt,
    middlelinewidth             =1pt,
    innermargin                 =0.3cm,
    outermargin                 =0.3cm,
    innerleftmargin             =0.3cm,
    innerrightmargin            =0.3cm,
    innertopmargin              =0.5cm,
    innerbottommargin           =0.5cm
}

\tcbuselibrary{skins}
\usetikzlibrary{shadings}

\colorlet{colexam}{red!75!black}

\usepackage{amsmath}
\usepackage{amssymb}
\usepackage{bm}
\usepackage{relsize}
\usepackage{siunitx}
\usepackage[super]{nth}

\usepackage{booktabs}
\usepackage{multirow}
\usepackage{colortbl}
\usepackage{makecell}
\usepackage{rotating}
\usepackage{threeparttable}
\usepackage{adjustbox}

\usepackage{float}
\usepackage{graphicx}
\usepackage{xcolor}

\usepackage{listingsutf8}

\usepackage{xparse}
\usepackage{xspace}

\usepackage[xindy,acronym]{glossaries}
\glsdisablehyper

\usepackage[hidelinks,bookmarks=false]{hyperref}

\usepackage[capitalise]{cleveref}

\definecolor{mycustomcolor}{HTML}{6D8764}

\tcbset{
  base/.style={
    enhanced,
    colframe=teal!5, 
    colback=teal!5,  
    sharp corners,
    borderline west={2pt}{0pt}{teal!70},
    boxrule=0.5mm,
    boxsep=4pt,
    leftrule=2mm,
    titlerule=0pt,
    fonttitle=\bfseries\itshape, 
    width=\textwidth, 
    attach boxed title to top left={
      xshift=2mm,
      yshift=-2mm
    }, 
    boxed title style={
      colback=teal!90, 
      colframe=teal!90, 
      sharp corners
    }
  }
}

\newtcolorbox{myexamplec}[2][]{
  base,
  title={#2}, 
}

\newcommand{\ra}[1]{\renewcommand{\arraystretch}{#1}}

\usepackage{caption}
\usepackage{subcaption}
\usepackage{pifont}

\title{Digital Twin-based Out-of-Distribution Detection in Autonomous Vessels}

\author{
  Erblin Isaku \\
  Simula Research Laboratory and \\
  University of Oslo \\
  Oslo, Norway \\
  \texttt{erblin@simula.no}
  \And
  Hassan Sartaj \\
  Simula Research Laboratory \\
  Oslo, Norway \\
  \texttt{hassan@simula.no}
  \And
  Shaukat Ali \\
  Simula Research Laboratory \\
  Oslo, Norway \\
  \texttt{shaukat@simula.no}
}

\begin{document}
\maketitle

\newcommand{\approach}{{\fontfamily{qhv}\selectfont ODDIT}}
\newcommand{\vda}{$\hat{A}_{12}$}

\begin{abstract}
    An autonomous vessel (AV) is a complex cyber-physical system (CPS) with software enabling many key functionalities, e.g., navigation software enables an AV to autonomously or semi-autonomously follow a path to its destination. Digital twins of such AVs enable advanced functionalities such as running what-if scenarios, performing predictive maintenance, and enabling fault diagnosis. Due to technological improvements, real-time analyses using continuous data from vessels' real-time operations have become increasingly possible. However, the literature has little explored developing advanced analyses in real-time data in AVs with digital twins built with machine learning techniques. To this end, we present a novel digital twin-based approach (\approach{}) to detect future out-of-distribution (OOD) states of an AV before reaching them, enabling proactive intervention. 
    Such states may indicate anomalies requiring attention (e.g., manual correction by the ship master) and assist testers in scenario-centered testing. The digital twin consists of two machine-learning models predicting future vessel states and whether the predicted state will be OOD. We evaluated \approach{} with five vessels across waypoint and zigzag maneuvering under simulated conditions, including sensor and actuator noise and environmental disturbances i.e., ocean current. \approach{} achieved high accuracy in detecting OOD states, with AUROC and TNR@TPR95 scores reaching 99\% across multiple vessels.
\end{abstract}

\keywords{Autonomous Vessels, Digital Twins, OOD Detection, SIL/HIL Testing}

\section{Introduction}

A digital twin of a cyber-physical system (CPS) aims to faithfully replicate the CPS, its environment, and its communication network to provide advanced capabilities (e.g., predicting unexpected behaviors in real-time before they occur) while considering the most up-to-date state of the CPS. The successful applications of digital twins for various CPSs have been demonstrated in various domains such as water treatment plants \cite{xu2021digital,CLDT}, train control and management systems \cite{KDDT}, industrial elevators \cite{ElevatorDT}, medical devices~\cite{sartaj2024modelbased,sartaj2024medet,sartaj2024uncertainty}, autonomous cars \cite{Time2EventDT}, and autonomous vessels (AVs) \cite{PredictiveDT2, PredictiveDTAVs}, to support advanced capabilities in real-time such as anomaly detection, time-to-event detection, supporting large-scale testing, and fault diagnosis. 

Autonomous Vessels (AVs) are cyber-physical systems responsible for the safe transit of passengers in a timely fashion while maintaining comfort as much as possible. AVs are complex systems comprising diverse hardware (e.g., sensors and actuators), control software, and a complex operating environment (i.e., ocean). The environment of AVs, like any other CPSs, is exposed to uncertainties that could affect the trajectory of the AVs, affecting the comfort of the passengers or, in the worst case, affecting the safety of the AVs. One key desired capability of AVs during their operation is to identify out-of-distribution states before an AV reaches those states. Out-of-distribution is a kind of anomaly that results in AVs not following a typical distribution of its parameters i.e., vessel controls and motions. Detecting such out-of-distribution before it happens can help AVs take necessary actions to take care of it either autonomously or with the help of ship masters. In addition, out-of-distribution detection enables software testers to focus on specific, high-risk scenarios by identifying the most critical situations that fall outside normal data distributions. 

Several approaches have been employed in the literature to detect out-of-distributions, such as~\cite{yang2021generalized, banerjee2024building}. However, these approaches run analysis during simulations to detect such out-of-distribution and improve the driving algorithm before deployment. On the other hand, enhanced hardware capabilities, like sensor deployment on AVs, now allow real-time data collection to identify out-of-distribution events before they occur.
To this end, we explore using digital twins to detect out-of-distribution in real-time. There have been works for digital twin-based analyses of AVs for fault diagnosis \cite{PredictiveDT2, PredictiveDTAVs} and path planning \cite{DTPathPlanning}. In the context of these works, a digital twin is built with models created by domain experts.

Compared to these works, we employ a data-driven approach (\approach{}) to build a digital twin of AVs to support real-time out-of-distribution detection. Moreover, we support out-of-distribution detection capability with this digital twin for AVs---a digital twin capability that has not been studied in the literature. The digital twin has two main components. First, a digital twin model that replicates the behavior of an AV, i.e., given the current state of the AV and controls (e.g., rudder and propeller), predicts the future state of the AV.  This component is built with Recurrent Neural Networks (RNN). The second component is the capability of the twin, that is, predicting out-of-distribution on the predicted future states of the AVs. These two components work together to perform out-of-distribution detection ahead of time. 
To evaluate \approach{}, we conducted experiments using five vessel models (i.e., \textit{Mariner}, \textit{Container}, \textit{Remus 100}, \textit{NPS AUV}, and \textit{Otter}), each with unique characteristics. We tested the models across two types of maneuvers---waypoint navigation and zigzag maneuvers---representing standard navigational patterns in AV operations and introducing varying degrees of complexity and control adjustments. 
Each vessel was subjected to a range of simulated disturbances to assess \approach{}'s effectiveness in detecting OOD events under real-world-like operational challenges. These disturbances include \begin{inparaenum}
    \item sensor noise, simulating erroneous reading or malfunctioning sensors, 
    \item actuator noise, represented by shifts in rudder angles, which emulate potential issues in steering control, and
    \item environmental disturbances, such as sudden spikes in ocean currents, which can unpredictably affect the vessel's stability and path.
\end{inparaenum}
Our results demonstrated that \approach{} consistently achieved high accuracy in detecting OOD states across these scenarios. Specifically, we observed that \approach{} achieved AUROC and TNR@TPR95 scores up to 99\%, underscoring its reliability in distinguishing normal operational states from potential anomalies, making it a highly effective tool for safety-critical applications.
Furthermore, we compare \approach{} with two alternative methods: an output-based approach and a distance-based approach. Our results show that \approach{} consistently outperforms both methods in OOD detection across sensor noise, actuator noise, and environmental disturbances for most vessels.

Our contributions are as follows: 
\begin{inparaenum}
    \item We introduce \approach{}, a novel digital twin framework for autonomous vessels that provides predictive capability for out-of-distribution detection in time series data. \item \approach{} employs a dual-component architecture: 
    \begin{inparaenum}
        \item a recurrent neural network that continuously predicts future vessel states based on historical and real-time data, and 
        \item a deep autoencoder model that evaluates these predicted states against normal operational behavior. This layered structure enables \approach{} to identify and respond to OOD events before they impact the AV.
        \end{inparaenum}
    \item Experimental results on multiple vessel models under diverse and challenging maneuvering conditions to demonstrate the effectiveness of our proposed approach---enabling proactive response capabilities that benefit both operational safety and testing.
\end{inparaenum}

\section{Background}\label{sec:background}
\subsection{Digital Twin}

\begin{figure}[tbp]
\vspace{0.3em}
  \centering
    \includegraphics[width=0.7\columnwidth]{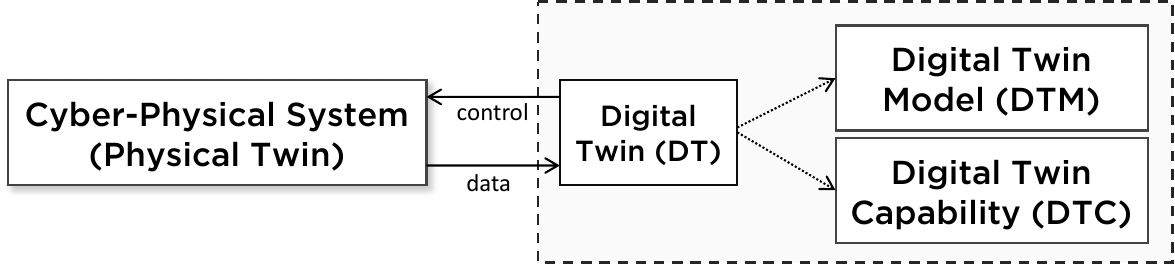}
    \caption{A conceptual model for digital twins for cyber-physical systems. }
    \label{fig:dtconceptualmodel}
\end{figure}

A digital twin is a virtual representation of a physical system that mirrors its real-time state, behavior, and processes through data collected from sensors and other sources, followed by providing advanced analyses~\cite{el2018digital}. The literature has many definitions and conceptual frameworks for digital twins~\cite{somers2023digital}. However, in our context, we adopt an existing conceptual framework for digital twins for cyber-physical systems (CPSs) \cite{xu2021digital,DTConceptual}. This conceptual framework has been successfully implemented to build data-driven digital twins in many domains, such as water treatment plants \cite{xu2021digital,CLDT}, train control and management systems \cite{KDDT}, industrial elevators \cite{Time2EventDT,ElevatorDT}, and autonomous driving \cite{Time2EventDT}. The conceptual model is shown in \cref{fig:dtconceptualmodel}. A CPS, a physical twin, is a system whose digital twin (DT) shall be built. DT has two essential parts: the digital twin model (DTM) and its capability (DTC). While the DTM represents the CPS using data-driven models such as machine learning (ML) to capture its behavior, the DTC defines the functionalities of the DT. These functionalities can include monitoring, predictive analytics (e.g., predicting failures before they actually happen in the system), uncertainty estimation/detection, and, in our case, out-of-distribution detection. The DTC processes real-time data from the physical twin to perform its capability. By interacting with the DTM, the DT performs its capability more efficiently and provides feedback to adjust the CPS operations, thereby preventing potential failures or unsafe conditions.

\subsection{Autonomous Vessels}
Maritime vessels are increasingly becoming autonomous, performing more operations without human intervention. In this paper, we refer to these as autonomous vessels (AVs), recognizing that autonomy levels vary. Det Norske Veritas (DNV)\footnote{\url{https://www.dnv.com/maritime/autonomous-remotely-operated-ships/}}, an international classification society, defines four AV autonomy levels: 1) remotely controlled vessels, operated from a distance; 2) vessels with onboard decision support, providing action recommendations to operators; 3) supervised autonomy, where vessels can take limited actions independently; and 4) full autonomy, with minimal operator interaction, except in exceptional situations. Our approach applies to AVs at any autonomy level, provided sufficient operational data is available.

AVs are complex cyber-physical systems composed of various subsystems, such as navigation (GPS-based path following), communication, control, and numerous sensors (e.g., for environmental monitoring and collision detection). A key functionality of AVs is path planning, enabling travel from a source to a target destination through different maneuvers. One common approach is waypoint-based navigation, where AVs follow a predefined path via guidance algorithms to reach their destination. Other maneuvers include zigzag and random paths. During maneuvering, each vessel is characterized by its degrees of freedom, representing motion along axes (i.e., x, y, and z) as well as rotational motion around those axes. In the maritime domain, those motions include surge, sway, and yaw, and, in 6DoF, additional motions like pitch, roll, and heave. 

AVs are prone to many uncertainties, such as errors in their sensor readings and environmental conditions (e.g., ocean currents and wind speed). Consequently, such uncertainties affect the AV's operation and potentially result in its states being out of distribution, which is abnormal and could lead to an unsafe situation. During the design and development of AVs, simulations are performed in different setups, such as Software in the Loop (SiL) and Hardware in the Loop (HiL). For SiL, Marine Systems Simulator (MSS)~\cite{MSSPaper,MSSGit} is commonly employed, which provides different vessel models, various maneuvers (e.g., zigzag), and the capability to simulate different environmental conditions (e.g., ocean currents).    

\section{Related Work}\label{sec:relatedwork}
\paragraph{Out-of-Distribution Detection}
Early out-of-distribution approaches include softmax-based detection in classification tasks, as demonstrated by Hendrycks and Gimpel~\cite{hendrycks2016baseline}, which showed its effectiveness in distinguishing in-distribution from OOD samples. Lee et al.~\cite{lee2017training} advanced this by using Generative Adversarial Networks (GANs) to generate synthetic OOD examples, improving classifier robustness through divergence minimization. In addition, distance-based methods such as Euclidean and Mahalanobis are commonly applied to enhance OOD detection by leveraging probability density, often combined with ensemble methods for improved performance \cite{lee2018simple, gwon2023out}. Other OOD methods include, output-based methods~\cite{liang2017enhancing}, outlier exposure~\cite{lu2023uncertainty}, gradient-based methods~\cite{igoe2022useful}, bayesian models~\cite{gal2016dropout}, OOD for foundation models~\cite{hendrycks2020pretrained}, density-based methods~\cite{ren2019likelihood}, theoretical analysis~\cite{jain2014multi}, reconstruction-based methods~\cite{zhou2022rethinking}, and hybrid models (e.g., a combination of reconstruction-based methods with distance-based methods)~\cite{denouden2018improving, daxberger2019bayesian}. 
While most OOD detection research focuses on image data, addressing distributional shifts in time series remains relatively unexplored. Recent work has explored system safety monitoring using probabilistic time-series forecasting to detect deviations in learned components~\cite{sharifi2024system}. In addition, autoencoder-based approaches have been utilized for OOD detection in black-box systems. For instance, SelfOracle~\cite{stocco2020misbehaviour} leverages an autoencoder combined with time-series anomaly detection to reconstruct input images and detect OOD instances based on reconstruction loss. Similarly, several methods employ variational autoencoders (VAEs) to quantify anomaly scores for failure prediction~\cite{hussain2022deepguard,borg2023ergo}. DeepGuard~\cite{hussain2022deepguard} applies VAE-based reconstruction errors to enhance vehicle safety by preventing roadside collisions, while Borg et al.~\cite{borg2023ergo} integrate VAEs with object detection to develop an OOD-aware emergency braking system. 

In time-series OOD detection, recent work by Banerjee et al.~\cite{banerjee2024building} closely relates to ours. It uses a quick change-point detection approach to monitor prediction errors in trajectory predictions for autonomous vehicles. Using statistical techniques like the Cumulative Sum (CUSUM), their method detects OOD events in real-time when deviations from expected behavior occur. While both approaches emphasize real-time OOD detection for safety, theirs rely on statistical techniques for sequential change detection in land-based vehicle trajectory data, whereas our method employs a machine learning-based approach integrating RNNs and autoencoders within a digital twin framework, tailored to the complex dynamics of autonomous vessel navigation in marine environments.

\paragraph{Digital Twins for Autonomous Vessels}
Digital twins have also been used in the context of autonomous vessels for various analyses. For instance, 
Hasan et al.~\cite{PredictiveDTAVs,PredictiveDT2} used predictive digital twins for state and parameter estimation of autonomous vessels to support fault diagnosis. Such a digital twin is built based on a graphical model of the autonomous vessel, and an adaptive Kalman filter is used for prediction. Using the same predictive digital twin approach, Hasan et al. \cite{FaultDiagnosisDT} also developed a real-time visualization of faults. Within the context of autonomous vessels, specifically the ones operating on the ocean surface, an application framework is proposed by Raza et al. \cite{DTApplicationFramework}. The framework proposes a layered architecture to demonstrate the integration of digital twins. The framework was validated with the 3D model of an autonomous vessel as the digital twin. Broadly speaking, digital twins have been explored for autonomous vessels for fault diagnosis \cite{PredictiveDT2,PredictiveDTAVs} and path planning \cite{DTPathPlanning}. Our work distinguishes itself from these works in the following directions. First, we focus on a new type of analysis with digital twins, i.e., out-of-distribution detection, which hasn't been performed with digital twins in the context of autonomous vessels. Second, we propose a new way of building digital twins based on data. Such a way of building a digital twin to support OOD detection in real-time is novel.  

\paragraph{Digital Twins of Cyber-Physical Systems}
Researchers have increasingly realized the importance of building digital twins of CPSs mainly because of their key advantages in advanced analyses such as predictive maintenance, anomaly detection, and what-if analyses \cite{DTConceptual}. For instance, digital twins have been used during the design of CPSs to assess whether they can handle security attacks \cite{eckhart_securing_2018}. Similarly, digital twins were also employed to assess the privacy of smart car systems---an example, CPS, to ensure whether such systems do not violate privacy requirements. Moreover, DTs have been applied in different CPS domains such as trains \cite{KDDT}, vertical transportation \cite{ElevatorDT,Time2EventDT}, various CPS testbeds \cite{CLDT}, and autonomous cars \cite{Time2EventDT}. These works demonstrate the importance of using digital twins to enable CPSs to perform analyses that couldn't be performed, for instance, directly by the CPSs. In contrast to these works, we investigate the use of digital twins in the context of autonomous vessels---a different domain to enable them to detect out-of-distribution in their parameters during operation to prevent such out-of-distribution in their parameters from happening. Moreover, our novelty lies in the fact that we build a data-driven digital twin to support real-time analyses, an aspect that is understudied in the domain of autonomous vessels.         

\section{Approach}\label{sec:approach}
In this section, we first provide an overview of the proposed approach. Next, we define the relevant terminologies to illustrate the approach. This is followed by a detailed discussion of each phase of the approach.

\subsection{\approach{} Overview}
\begin{figure}[tbp]
\vspace{0.3em}
  \centering
    \includegraphics[width=\columnwidth]{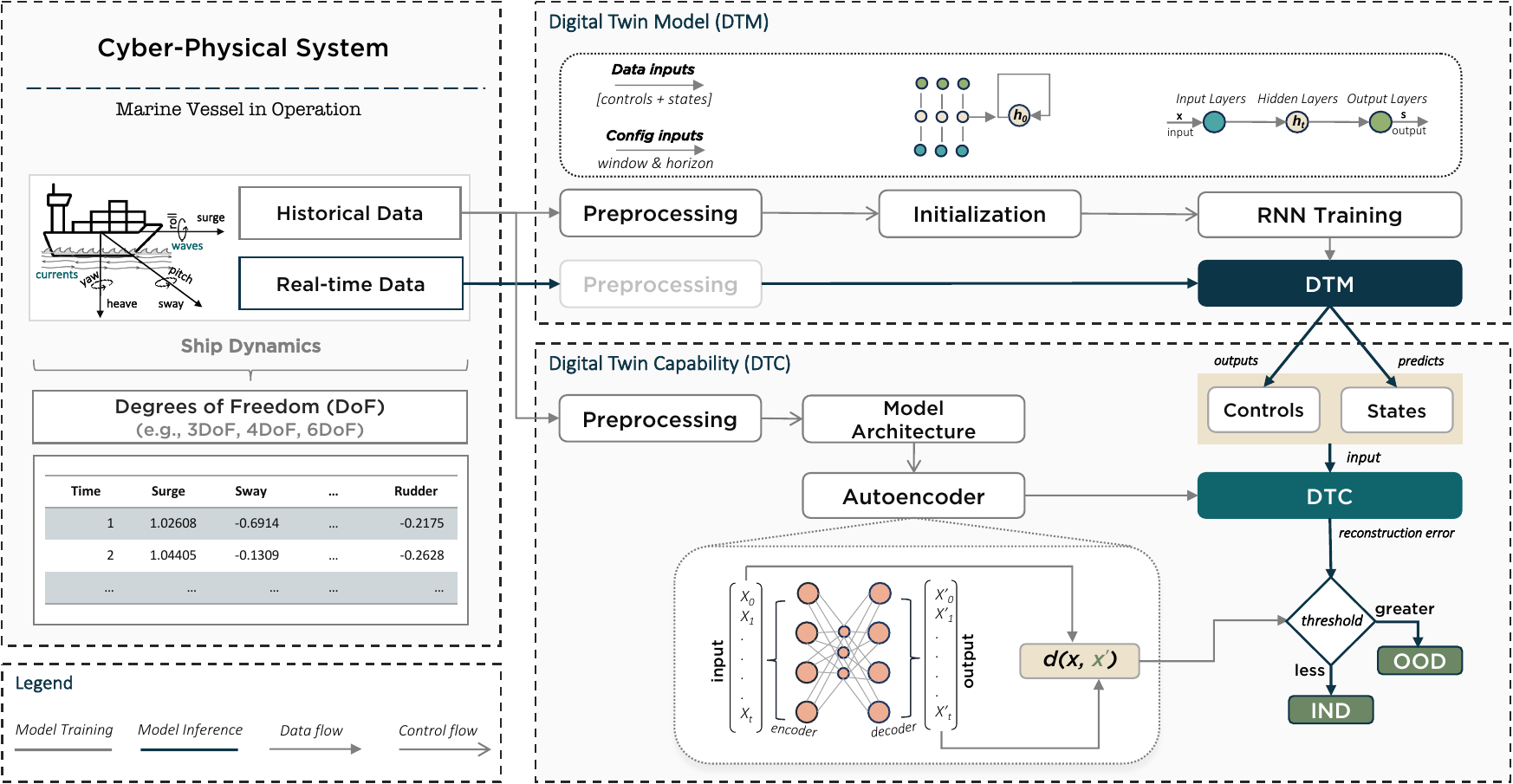}
    \caption{Overview of \approach{}, highlighting the DT and the corresponding physical system.}
    \label{fig:ood-framework}
\end{figure}

The overall framework of \approach{} is depicted in \cref{fig:ood-framework}. 
\approach{} accepts two types of inputs: data and configurations. 
Data inputs constitute historical data collected during AV's normal operations, incorporating vehicle control and state information. 
Configuration inputs consist of two key parameters: a \emph{window}, which specifies the number of past states to use during training, and a \emph{horizon}, which defines the number of future states to predict. 
\approach{} has three phases: (i) building DTM, (ii) creating DTC, and (iii) operating DT with real-time data from the marine vessel during either simulation or actual operation. 
During the first phase, we build the DTM utilizing historical data representing the in-distribution dataset. 
This process involves preprocessing the in-distribution dataset, initializing the RNN, and subsequently training the RNN to build the DTM. 
In the second phase, we create the DTC using the historical in-distribution dataset. 
This involves preprocessing the data, outlining the model architecture, and training the Autoencoder.  
The trained Autoencoder then functions as the DTC during the operation of the DT. 
In the third phase, we integrate DTM and DTC to form a unified DT that operates alongside its physical counterpart for OOD prediction. 
During DT operation, real-time data from the marine vessel is continuously streamed to the DT. 
The DTM then predicts the next states, as determined by the configuration parameter, i.e., \emph{horizon}.
These predicted states and the real-time data are then provided to DTC, which employs Autoencoder to determine potential in/out-of-distribution scenarios.
It is important to note that both RNN and Autoencoder are trained with in-distribution data, any real-time data point deviating from learned in-distribution/expected behavior implies OOD.


\subsection{Definitions}

\noindent\textbf{Definition 1 (State).} A state is defined as \mbox{$S = \{s_1, s_2, \dots, s_n\}$}, where $n$ is the total number of state elements and each $s_x$ represents a particular vehicle motion parameter, such as surge velocity, roll rate, and pitch angle. 

\noindent\textbf{Definition 2 (Next States).} A set of next/predicted states is defined as \mbox{$S_n = \{S^{t_1}, S^{t_2}, \dots, S^{t_h}\}$}, where $S$ represents a state at a specific time step with a maximum time horizon of $h$. 

\noindent\textbf{Definition 3 (Control).} A vehicle's control is defined as \mbox{$C = \{C_1, C_2, \dots, C_m\}$}, where $m$ is a total number of control parameters supported by a vehicle and $C_x$ denotes a particular control parameter such as ocean current speed, rudder angle, and propeller. 

\subsection{Data Collection}\label{ssec:datacollection}
Our approach relies on historical in-distribution data representing AV's normal behavior. 
This data is primarily collected either from simulations or during actual marine missions. 
For autonomous systems, a common practice is to collect data during simulation via software-in-the-loop (SIL) or hardware-in-the-loop (HIL) simulators~\cite{sartaj2021testing,sartaj2021automated}. 
This process involves defining various scenarios that include specific maneuvers or paths for the target AV, considering the AV's controls and DoF.
These scenarios are typically formulated as simulator scripts, which are loaded into a SIL/HIL simulator to simulate the AV's behavior in autonomous mode. 
An alternative method involves a semi-autonomous or guided mode, in which the ship master takes control of the AV in response to environmental disturbances. 
In either of the simulation methods, the AV's data is recorded at each time step.
In our approach, we first need to define scenarios for the targeted AV, incorporating maneuvers that the AV can perform. 
These maneuvers can represent various paths, such as zigzag or circular routes, depending on the controls and DoF supported by the AV. 
The next step is to use a vessel simulator to simulate the AV behavior, running in either autopilot (autonomous) or guided (semi-autonomous) mode. 
During the simulation, we record AV's data, including the control ($C$) and state ($S$) information, after a one-second time step. 
It is important to note that such data can also be gathered from the real operations of the AV. 
When formulating scenarios, it is essential to ensure that the overall duration of each simulation scenario is adequate for collecting sufficient data for training purposes. 
Moreover, the AV data collected during this process does not require labeling.
Our approach utilizes unlabeled data for building DTs.

\subsection{Determining Model Settings}\label{ssec:modelsetting}
Our approach employs ML models, including RNN and Autoencoders. 
These models were chosen based on their alignment with the specific requirements of time-series forecasting and OOD detection, as well as their effectiveness in similar contexts~\cite{baier2023relinet, nicholaus2021anomaly}.
While our study did not exhaustively evaluate all possible model types, we conducted preliminary experiments comparing RNNs and autoencoders to alternative approaches, such as Long Short-Term Memory (LSTM), Gated Recurrent Units (GRU), and Variational Autoencoders (VAEs).
Our findings suggested that RNNs and autoencoders offered a more effective balance between predictive accuracy, computational efficiency, and suitability for real-time applications.
Nonetheless, further comprehensive evaluations are necessary to fully establish the optimal models.

For effective model training and application in a specific context, it is necessary to determine the optimal architecture and parameters. 
This involves choosing the most suitable model structure and tuning the relevant parameters to ensure the model's performance for the specific application. 
For the RNN, we used the ReLiNet framework~\cite{baier2023relinet}, which has an inbuilt mechanism for determining optimal model settings. 
For the Autoencoder, we conducted a pilot study to identify suitable configurations. 
Specifically, we used Optuna~\cite{akiba2019optuna}, a widely used optimization framework. 
Optuna requires specifying optimization objectives, which include different model and hyperparameter settings. 
Initially, we defined model architectures containing different layers, varying dimensions within each layer, and various activation functions such as ReLU and Sigmoid. 
Subsequently, we specified different learning rates, batch sizes, optimizers, and loss functions. 
After setting the optimization objectives, we executed the experiment for 100 epochs and 50 trials. 
To assess the optimization objectives, we used the Area Under the Receiver Operating Characteristic Curve (ROC AUC) as our evaluation metric. 
At the end of the experiment execution, we selected the model architecture and hyperparameters that provided the best results according to Optuna's suggestions.

The architecture of the RNN model comprises an input layer, two recurrent layers, each with a dimension of 256, an output layer, and a ReLU activation function. 
The Autoencoder model consists of an input layer, three linear layers, an output layer, and RReLU activation functions for the inner layers and a Sigmoid activation function for the output layer.
In the Autoencoder model, the encoder segment consists of layers with dimensions of 64, 32, 16, and 8. 
Moreover, the decoder segment comprises layers with dimensions of 8, 16, 32, and 64. 
The remaining hyperparameters for RNN and Autoencoder are presented in \cref{tab:params}. 
Some of the parameter values are common across vessels, however, some are specific to a particular vessel type.


\begin{table}[!t]
    \centering
    \scriptsize
    \noindent
    \caption{Hyperparameter values for RNN and Autoencoder}
    \resizebox{0.7\textwidth}{!}{%

    \begin{tabular}{p{.15\textwidth}p{.13\textwidth}p{.18\textwidth}}
        \toprule
        \multicolumn{1}{l}{Parameter} & RNN & Autoencoder \\
        \cmidrule(lr){1-1}\cmidrule(lr){2-2}\cmidrule(ll){3-3}
        Learning rates & 0.001, 0.0025 & 0.0002, 0.001, 0.002, 0.007 \\
        Batch sizes & 128, 256 & 16, 64, 70, 84 \\
        Epochs & 600, 1000 & 100 \\
        Loss function & {MSE} & {MSE} \\
        Optimizer & {Adam} & {Adam}\\
        \midrule
        Window & 60 & - \\
        Horizon & 60 & - \\
        \bottomrule
    \end{tabular}}
    \label{tab:params}
\end{table}

\subsection{Building DTM}
The primary objective of the DTM is to learn the behavior of an AV and predict its potential next states.
Given that the data from a specific AV at a certain time step ($t$) maintains a temporal relationship with the data at the previous time step ($t-1$), RNNs are considered appropriate for this scenario.
RNNs are a type of neural network specially designed to work with time series data.
They have the unique ability to remember past information through hidden layers and use this information for future predictions.
Moreover, different variants of RNN have been used for various autonomous systems, such as the LSTM model~\cite{wu2020anomaly,sartaj2024automated}. 
In our approach, we use ReLiNet~\cite{baier2023relinet}---a variant of RNN---specifically designed for multistep prediction in highly dynamic environments. 
ReLiNet has demonstrated its effectiveness over other RNN variants in the context of autonomous vessels~\cite{baier2023relinet}.

\subsubsection{Preprocessing}\label{ssec:preprocessing}
The raw data collected (\cref{ssec:datacollection}) for a specific AV captures the vessel's maneuvering motion, including both state variables (e.g., sway velocity, yaw rate) and control variables (e.g., rudder angle, propeller speed), based on its respective degrees of freedom.
Preprocessing such data is an initial step for training ML models.
For this purpose, we apply the widely used Min-Max normalization technique.
By employing the Min-Max normalization technique, we transform the raw data values in the dataset to a consistent scale, specifically within the range of 0 to 1. 
After this step, we obtain the preprocessed data that is now prepared and suitable for training.  
This data is initially used for the pre-training/initialization process, followed by the training of the ReLiNet.

\subsubsection{Initialization and Training}
The full training process of ReLiNet comprises two parts: an initialization or pre-training phase, which is designed to initialize hidden states of the model, and a training phase, which is designed to train the model. 
The model's architecture and hyperparameters, utilized for initialization and training, are determined based on a pilot study, as detailed in \cref{ssec:modelsetting}.
For executing the training process, we specify 600 epochs for initialization and 1000 for training (\cref{tab:params}). 
The training process starts with the initialization phase. 
After a successful initialization of hidden states, the actual training process begins. 
Upon completion of this process, the trained ReLiNet model is preserved. 
This model serves as the DTM of the AV for predicting its next states.

\subsection{Creating DTC}\label{ssec:dtc}
The primary objective of the DTC is to enhance the DT with predictive capabilities, allowing it to determine whether an AV is adhering to the expected maneuverability (i.e., within distribution) or deviating from it (i.e., out of distribution) during its operation.
Such deviations could be caused by malfunctions in sensors and actuators or environmental disturbances. 
Identifying such deviations (i.e., OOD) is essential for the safe operation of an AV. 
In our approach, we construct a DT capable of detecting OOD using an Autoencoder.
Autoencoders are extensively used for anomaly and OOD detection in ML models~\cite{yang2021generalized, jiang2023read, zhou2022rethinking}. 
Therefore, we employ an Autoencoder in our context to create the DTC. 
It is important to note that while anomaly detection pertains to identifying system faults or failures, OOD detection aims to detect deviations from expected behavior potentially caused by system failure or environmental factors.

To create the DTC, we first preprocess the historical in-distribution data following the process outlined in \cref{ssec:preprocessing}. 
Subsequently, we utilize the Autoencoder model architecture determined through a pilot study, described in \cref{ssec:modelsetting}. 
We then proceed to train the Autoencoder model using the processed data and the hyperparameters provided in \cref{tab:params}. 
During the training phase, we compute the reconstruction error ($RE_x$) for each predicted state using \cref{eq-re}. 
In this equation, $S^{t_i}\in S_n$ represents the predicted next state at the time step ($t_i$), $Y_i$ denotes the state reconstructed by the Autoencoder, and $k$ stands for the total number of predicted states.

\begin{equation}\label{eq-re}
RE_{x} = \sqrt{\sum_{i=1}^{k} (S^{t_i} - Y_i)^2}
\end{equation}

After calculating reconstruction errors for each of the predicted states in $S_n$, we compute the threshold ($T_{OOD}$) using \cref{eq-oodthreshold}. 
In this equation, $RE$ denotes the list of reconstruction errors computed corresponding to each predicted state, i.e., $RE = \{RE_1, RE_2, ..., RE_h\}$, where $h$ represents \emph{horizon}. 
The threshold ($T_{OOD}$) is calculated as the mean of the training reconstruction errors, plus three times their standard deviation, following the common practice~\cite{pukelsheim1994three}. 
This implies that any data point with a reconstruction error greater than this threshold would be considered out-of-distribution. 
At the end of the training process, we store the trained Autoencoder model. 
This model now represents the AV's DTC, which we employ to predict the AV's OOD behavior during its operation alongside DT. 
Furthermore, we keep the threshold ($T_{OOD}$) value, which is employed to determine whether a specific data point falls within or outside of the distribution (IND/OOD) while operating the DT.

\begin{equation}\label{eq-oodthreshold}
T_{OOD} = \mu(RE)+3*\sigma(RE)
\end{equation}

Our approach intentionally sets the threshold as the mean of the reconstruction errors plus three times their standard deviation, prioritizing the reduction of false negatives (missed OOD detections) even at the cost of a slight increase in false positives (misclassified IND samples as OOD).
This design choice aligns with practical operational scenarios, where the consequences of failing to detect OOD states—such as undetected anomalies or unsafe conditions—are more critical than the occasional misclassification of normal states.
Nevertheless, future work will systematically evaluate and refine this threshold-setting method to further optimize the trade-off between false positives and false negatives.

\subsection{Operating DT with AV}\label{ssec:operatedt}
To operate DT alongside AV, first, we load the trained DTM and DTC and prepare them for inference. 
During AV operation, the real-time data received from an AV is preprocessed using the same method elaborated in \cref{ssec:preprocessing}. 
We provide processed data containing the AV's control ($C$) and state ($S$) information along with \emph{window} and \emph{horizon} values to the DTM.
Based on the specified \emph{window} and \emph{horizon} values, DTM predicts the next possible states ($S_n$) of AV. 
For instance, if both \emph{window} and \emph{horizon} are 10, the DTM will use the past 10 states' information to predict the next 10 states.

For reconstructing states, we provide the output from DTM, i.e., $S_n$ and AV controls data ($C$) to DTC. 
Specifically, we use each predicted next state at each time step ($S^{t_i}\in S_n$) and reconstruct the state denoted as $Y_i$. 
Using the input state ($S^{t_i}$) and the reconstructed state ($Y_i$), we calculate the reconstruction error. 
For this purpose, we use \cref{eq-re} to calculate the reconstruction error, denoted as $RE_r$. 
The reconstruction error is a measure of the difference between the original input ($S^{t_i}$) and the reconstructed output ($Y_i$), which indicates a deviation for a single state.

To determine whether the input real-time data is IND or OOD, we utilize the threshold $T_{OOD}$, which is calculated using \cref{eq-oodthreshold} during the training phase (\cref{ssec:dtc}). 
For the real-time data point denoted as $D_r$ and the reconstruction error $RE_r$ corresponding to predicted states, we use \cref{eq-ood} to decide IND or OOD. 
According to this equation, if $RE_r$ is less or equal to the threshold $T_{OOD}$, the given data point ($D_r$) is IND. 
This indicates that the AV is following the expected path. 
On the other hand, if $RE_r$ is greater than the threshold $T_{OOD}$, this is considered potential OOD. 
This indicates that the AV is potentially deviating from the expected path. 

\begin{equation}\label{eq-ood}
  D_r = \left\{ 
  \begin{array}{ll}
    \text{IND} & \quad RE_r \leq T_{OOD} \\
    \text{OOD} & \quad RE_r > T_{OOD} 
  \end{array}
\right.
\end{equation}

\section{Empirical Evaluation}\label{sec:experiment}
We aim to assess \approach{}'s performance in detecting OOD occurrences during various AV maneuvers, specifically focusing on (i) sensor and actuator noise types, and (ii) environmental disturbances. 
In addition, we compare \approach{} with traditional methods for OOD detection. 
Considering these factors, we define the following research questions (RQs).

\begin{itemize}
    \item [\textbf{RQ1}] \emph{How effective is \approach{} in detecting OOD occurrences when the AV experiences sensor noise?}
    \item [\textbf{RQ2}] \emph{How effective is \approach{} in detecting OOD occurrences when the AV experiences actuator noise?}
    \item [\textbf{RQ3}] \emph{How accurately can \approach{} identify OOD occurrences when the AV is subjected to environmental disturbances?}
    \item [\textbf{RQ4}] \emph{How does the performance of \approach{} compare with traditional methods for OOD detection?}
\end{itemize}

\textbf{RQ1} and \textbf{RQ2} study \approach{}'s ability to detect OOD occurrences under sensor and actuator noise, respectively, by analyzing: \begin{inparaenum}[(a)]
    \item \textit{Vessel Types}, assessing detection consistency across different AV models;
    \item \textit{Noise Magnitudes}, evaluating the impact of increasing noise levels on performance;
    \item \textit{Vessel and Noise}, studying whether certain vessels are more or less affected by noise variations;
    \item \textit{Correlation Analysis}, investigating statistical relationships between noise intensity and detection effectiveness.
\end{inparaenum}
\textbf{RQ3} investigates \approach{}'s ability to detect OOD occurrences triggered by environmental disturbances (i.e., ocean currents)—unpredictable factors that can disrupt AV operations. We assess its effectiveness across different vessel types.
Lastly, \textbf{RQ4} focuses on analyzing the extent of improvement \approach{} provides over existing methods in detecting OOD. For this comparative analysis we consider \begin{inparaenum}[(a)]
    \item sensor noise,
    \item actuator noise, and
    \item environmental disturbances. 
\end{inparaenum}



\subsection{Vessel Selection}
For the experiments, we used MSS~\cite{MSSGit} to simulate vessels. 
The MSS, initially developed for education, is widely used in maritime simulations for dynamic positioning, station-keeping, and maneuvering under disturbances. Built on hydrodynamic principles, it serves both industry and academia. In industry, it aids in designing and validating control strategies for autonomous and remotely operated vessels~\cite{bo2015marine}. In academia, it supports research in state prediction, fault detection, and system identification~\cite{baier2021hybrid,baier2023relinet}. Its models are built on well-established hydrodynamic principles, making it a reliable platform for replicating complex marine operations and validating vessel control strategies~\cite{fossen2011handbook}.

Specifically, we selected a diverse set of vessel models for this study, which enabled us to evaluate the DT-based OOD detection framework across multiple operational and environmental scenarios. The vessel models were chosen to represent a wide range of motion and dynamics. For this, we focused on two key criteria: the types of maneuvers each vessel can support and its ability to handle environmental disturbances (e.g., wind and/or ocean current). 
We used the data from the following vessels, whose key characteristics are presented in~\cref{tab:vessel_characteristics}. 

\textbf{Mariner-Class Cargo Vessel} denoted as Mariner, is equipped with a rudder and maneuvers in a 3DoF. The vessel states are: \textit{surge velocity}, \textit{sway velocity}, \textit{yaw rate}, and \textit{yaw angle}. Mariner maneuvers in a 2D path-following (x and y coordinates) with a default of 6 waypoints. This vessel is used for both waypoint and zigzag maneuvering, and does not support ocean current as a disturbance. 

\textbf{Container Vessel} denoted as Container operates with a rudder control and maneuvers in 4DoF. The vessel states include: \textit{surge velocity}, \textit{sway velocity}, \textit{yaw rate}, \textit{yaw angle}, \textit{roll rate}, and \textit{roll angle}. This vessel supports zigzag maneuvering and includes no environmental disturbances (i.e., ocean currents). 

\textbf{Remus 100 Autonomous Underwater Vehicle (AUV)} denoted as Remus 100, is equipped to operate under depth while exposed to ocean currents. It maneuvers in 6DoF, with state variables including: \textit{surge velocity}, \textit{sway velocity}, \textit{heave velocity}, \textit{roll rate}, \textit{pitch rate}, \textit{yaw rate}, \textit{roll angle}, \textit{pitch angle}, and \textit{yaw angle}. This AUV follows a 3D path with specified waypoints (default 6) in x, y, and z coordinates, and supports both waypoint and zigzag maneuvers, responding to control inputs for rudder, stern-plane, and propeller speed to maintain stability in underwater environments. 

\textbf{Naval Postgraduate School (NPS) Autonomous Underwater Vehicle (AUV)}, denoted as NPS AUV, maneuvers in a 6DoF with states: \textit{surge velocity}, \textit{sway velocity}, \textit{heave velocity}, \textit{roll rate}, \textit{pitch rate}, \textit{yaw rate}, \textit{roll angle}, \textit{pitch angle}, and \textit{yaw angle}. The vehicle follows a 3D path with x, y, and z waypoints (default 7) under depth and heading control, and operates in the presence of ocean currents, responding to control inputs for the rudder, stern plane, port bow plane, starboard plane, and propeller speed. 

\textbf{Otter Uncrewed Surface Vehicle (USV)} denoted as Otter, is designed for path-following tasks under various control strategies, navigating in 6DoF. The Otter can operate in the presence of ocean currents and utilizes state variables including: \textit{surge velocity}, \textit{sway velocity}, \textit{heave velocity}, \textit{roll rate}, \textit{pitch rate}, \textit{yaw rate}, \textit{roll angle}, \textit{pitch angle}, and \textit{yaw angle}. It can follow a 2D path (north-east positions) through a series of waypoints (default 7) with coordinates in x and y. Control is achieved through differential thrust using left and right propellers, managed by strategies such as heading autopilot to maintain course.

\begin{table}[h!]
    \ra{1.3}
    \centering
    \caption{Vessel characteristics by DoF, motion controls, maneuvers, and environmental disturbances}
    \resizebox{\textwidth}{!}{%
    \Large
    \begin{tabular}{p{.15\textwidth} p{.08\textwidth} p{.35\textwidth} p{.25\textwidth} p{.20\textwidth} >{\centering\arraybackslash}p{.16\textwidth}}
        \specialrule{1.1pt}{1pt}{1pt}
        Vessel Name & DoF & Control Names & Motion & Maneuver & Ocean Current \\
        \cmidrule(lr){1-1} \cmidrule(lr){2-2} \cmidrule(lr){3-3} \cmidrule(lr){4-4} \cmidrule(lr){5-5} \cmidrule(lr){6-6}
        
        Mariner & 3DoF & Rudder Angle & Surge, Sway, Yaw & Waypoint \& Zigzag & \ding{55} \\
        
        Container & 4DoF & Rudder Angle & Surge, Sway, Yaw, Roll & Zigzag & \ding{55} \\
        
        Remus 100 & 6DoF & Rudder Angle, Stern Plane Angle, Propeller & Surge, Sway, Heave, Roll, Pitch, Yaw & Waypoint \& Zigzag & \ding{51} \\
        
        NPS AUV & 6DoF & Rudder Angle, Stern Plane Angle, Port Bow Plane Angle, Starboard Plane Angle, Propeller & Surge, Sway, Heave, Roll, Pitch, Yaw & Waypoint & \ding{51} \\

        Otter & 6DoF & Left Propeller, Right Propeller & Surge, Sway, Heave, Roll, Pitch, Yaw & Waypoint & \ding{51} \\
        
        \specialrule{1.1pt}{1pt}{1pt}
    \end{tabular}%
    }
    \label{tab:vessel_characteristics}
\end{table}

\paragraph{Data Collection}
The dataset generation process was specifically designed to capture unique attributes and operational dynamics of each vessel, as detailed in \cref{tab:vessel_characteristics}. The features used for the models include control parameters (e.g., rudder angle, propeller speed) and motion-related variables (e.g., surge velocity, sway velocity, yaw rate), capturing the vessel's behavior during different maneuvers, such as waypoint navigation and zigzag, an example shown in \cref{tab:dataset_example}. In scenarios where environmental conditions can be altered, we simulate more extreme values of ocean currents, creating an additional dataset.
Depending on the specific scenario, such as the type of maneuver or the ocean currents, varying settings, and conditions were applied for data collection. The detailed configurations and conditions for each scenario are further elaborated in the following section.


\renewcommand{\arraystretch}{1.5} 
\begin{table}[h!]
\centering
\caption{Example of collected data}
\resizebox{\textwidth}{!}{
\fontsize{12pt}{16pt}\selectfont
\begin{tabular}{|c|c|c|c|c|c|c|c|c|}
\hline
Time (s) & Surge Velocity (m/s) & Yaw Rate (rad/s) & Yaw Angle (rad) & ... & Rudder Angle (rad) & Propeller (rpm) & Ocean Current Speed (m/s) \\ \hline
0   & 1.0000  & 0.0000   & 1.5561   & ... & 0.0000     & 1000.0000   & 0.5005          \\ \hline
1   & 1.0264  & -0.0055  & 1.5522   & ... & -0.2181    & 1007.8646   & 0.4974          \\ \hline
2   & 1.0438  & -0.0035  & 1.5472   & ... & -0.2532    & 1015.7292   & 0.5050          \\ \hline
... & ...     & ...      & ...      & ... & ...         & ...         & ...             \\ \hline
3598 & 1.6334 & 0.0000   & 1.6904   & ... & 0.0001     & 1300.0339   & 0.5016          \\ \hline
3599 & 1.6334 & -0.0000  & 1.6904   & ... & -0.0001    & 1300.0339   & 0.4997          \\ \hline
3600 & 1.6334 & 0.0000   & 1.6904   & ... & 0.0001     & 1300.0339   & 0.4983          \\ \hline
\end{tabular}
}
\label{tab:dataset_example}
\end{table}

\subsection{Experimental Setup}
We implemented \approach{} using Python. 
To develop DTM, we used ReLiNet~\cite{baier2023relinet} to create the RNN model. 
For the implementation of DTC, we utilized the PyTorch deep learning library to create a deep Autoencoder model.
In the following, we discuss the specifications of vessel maneuvers and the corresponding setup for each RQ.

\subsubsection{Maneuver Specification}\label{ssec:maneuvers}
\begin{figure}[tbp]
  \centering
  \begin{subfigure}[t]{0.65\textwidth}
    \centering
    \begin{subfigure}[t]{0.48\textwidth} 
      \centering
      \includegraphics[width=\textwidth]{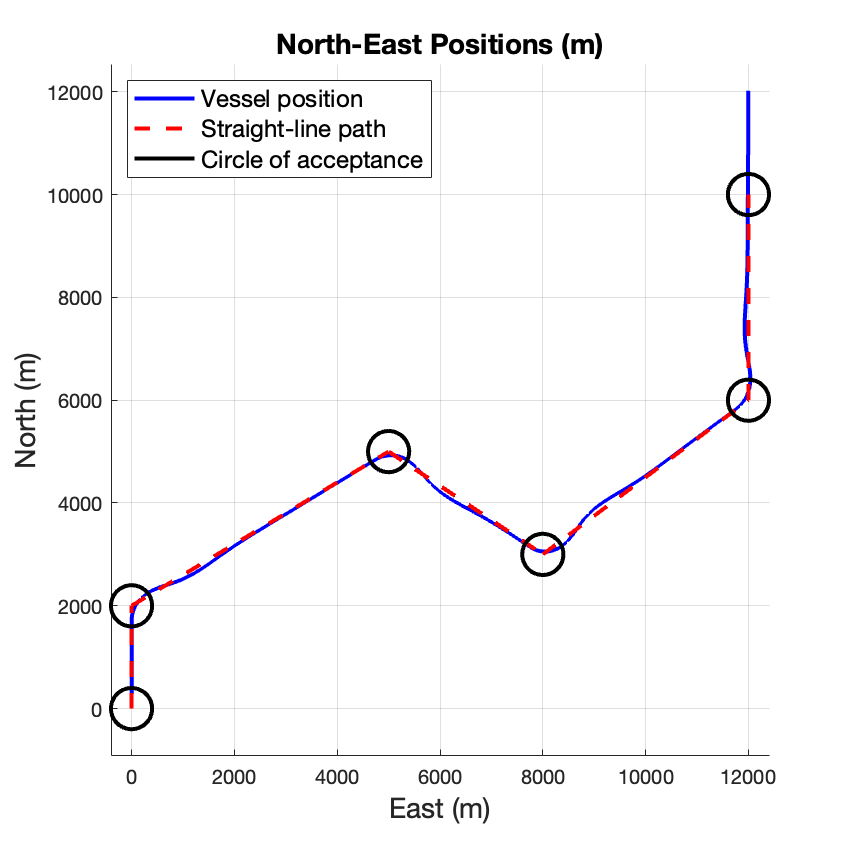}
      \caption{2D path following using an autopilot.}
      \label{fig:waypoint-following-ind}
    \end{subfigure}%
    \hfill
    \begin{subfigure}[t]{0.48\textwidth} 
      \centering
      \includegraphics[width=\textwidth]{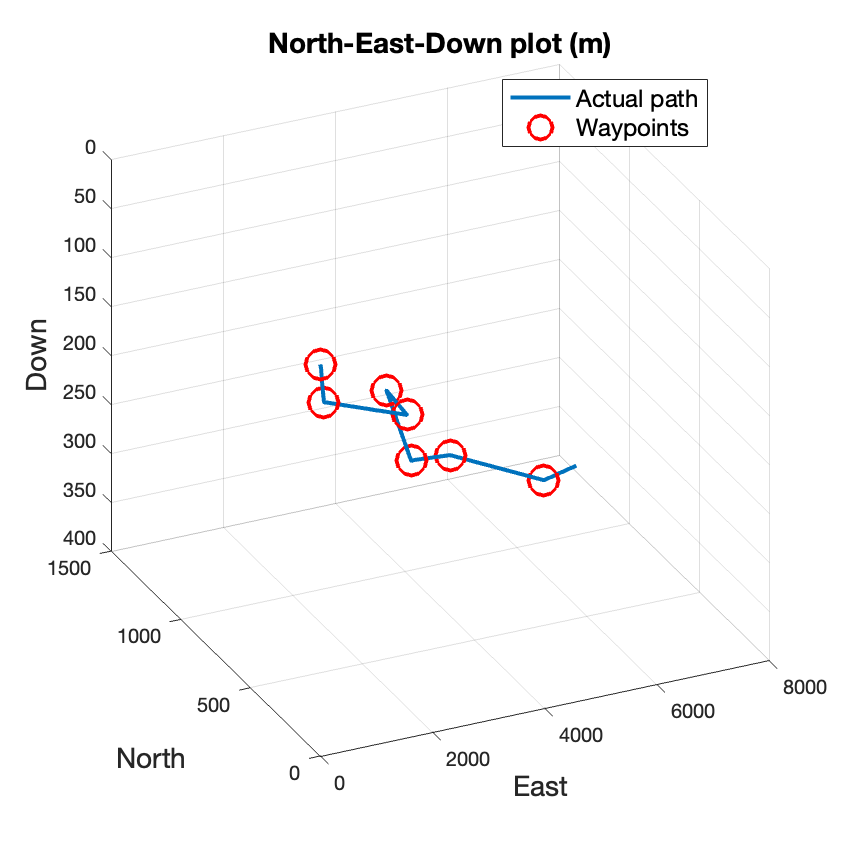}
      \caption{3D path following using an autopilot.}
      \label{fig:waypoint-following-3d}
    \end{subfigure}
  \end{subfigure}%
  \hfill
  \begin{subfigure}[t]{0.3\textwidth} 
    \centering
    \includegraphics[width=\textwidth]{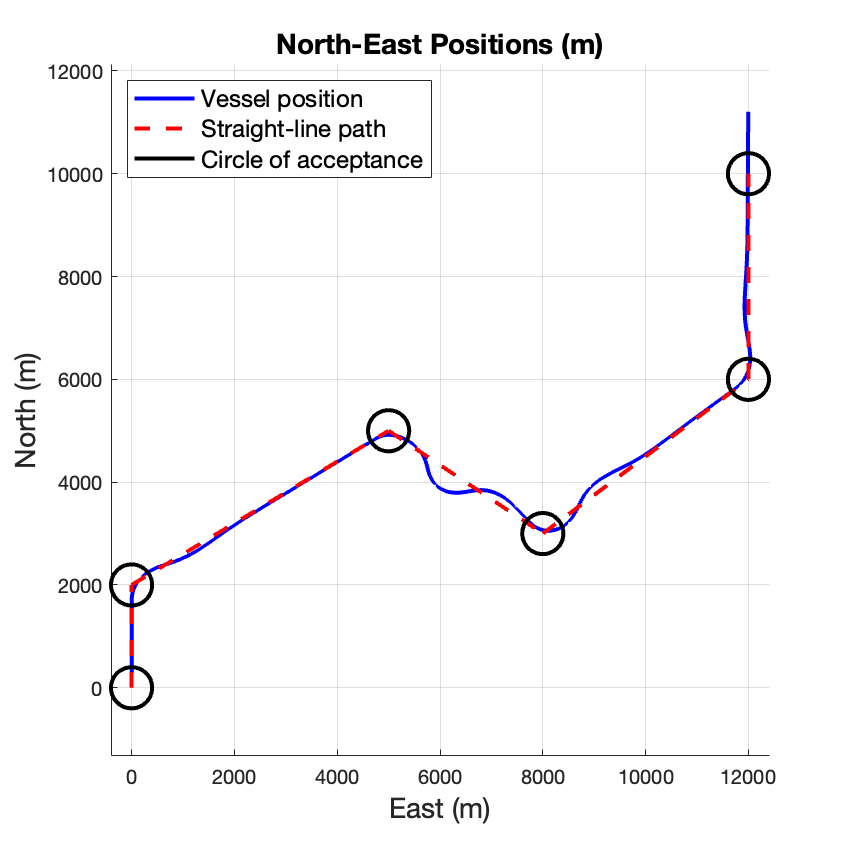}
    \caption{Vessel's OOD behavior due to the sensor noise.}
    \label{fig:waypoint-following-ood}
  \end{subfigure}
  \caption{Examples of waypoint navigation: (a) 3DoF vessel navigating in a 2D path, (b) 6DoF vessel navigating in a 3D path, and (c) OOD behavior due to noise in position sensors.}
  \label{fig:waypoint-following-combined}
\end{figure}

\begin{figure}[tbp]
  \centering
  \begin{minipage}[b]{0.45\columnwidth}
    \centering
    \includegraphics[width=\textwidth]{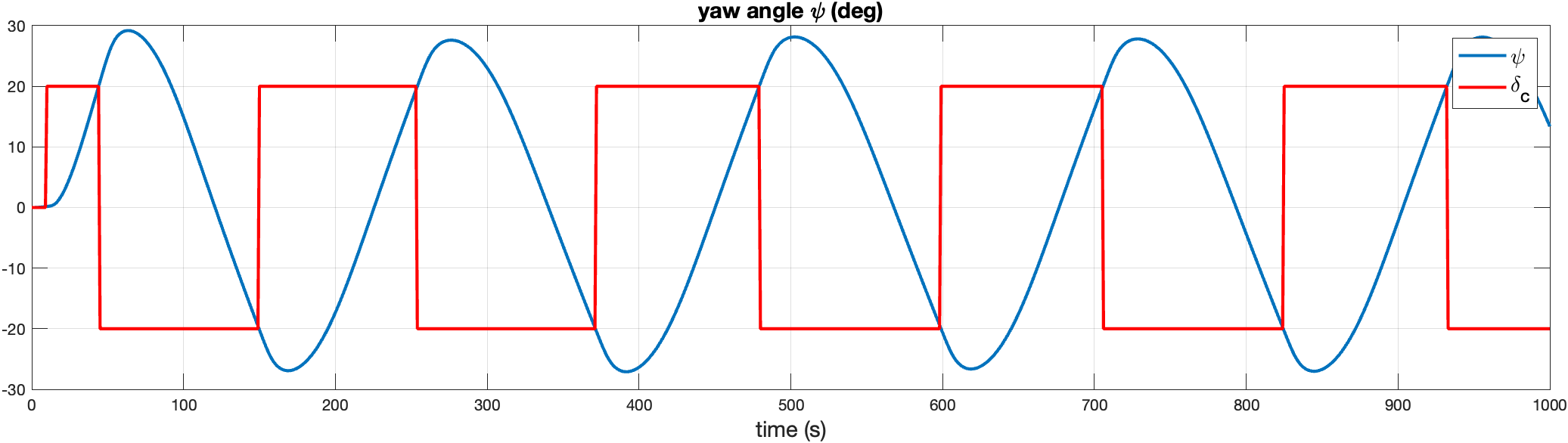}
    \caption{Zigzag with normal 20° rudder angle.}
    \label{fig:zigzag-maneuver-ind}
  \end{minipage}%
  \hfill 
  \begin{minipage}[b]{0.45\columnwidth}
    \centering
    \includegraphics[width=\textwidth]{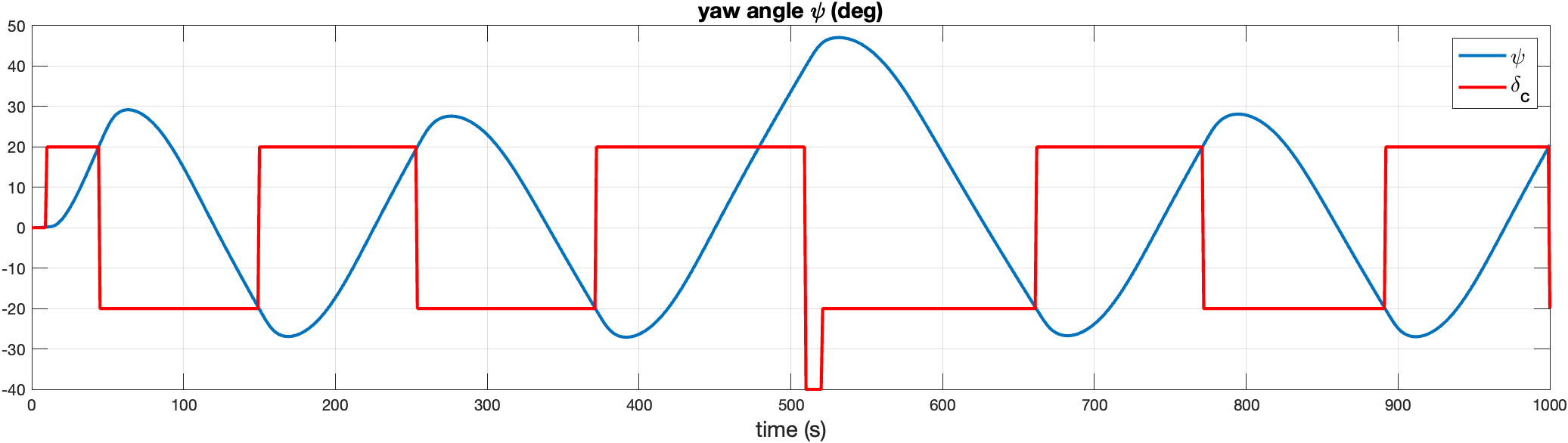}
    \caption{Zigzag with high 40° rudder angle.}
    \label{fig:zigzag-maneuver-ood}
  \end{minipage}
\end{figure}

For the experiments, we design two types of paths considering AV's operating modes, i.e., semi-autonomous (guided) mode and autonomous (autopilot) mode. 
For the semi-autonomous mode, we formulate paths characterized by a zigzag pattern, termed as \emph{zigzag maneuver}.
In the case of the autonomous mode, we create paths driven by waypoints, referred to as \emph{waypoint navigation}. 
It is important to note that while multiple types of paths are possible, testing AVs in all conceivable scenarios is impractical. 
Hence, testers typically devise specific paths that align with their testing objectives. 
For instance, zigzag maneuvers are commonly used to evaluate an AV's maneuverability, such as its responsiveness in sway, yaw, and other motions~\cite{yasukawa2015introduction, duman2017prediction}.
\cref{fig:zigzag-maneuver-ind} shows an example of a zigzag maneuver with a normal rudder angle of 20°. 
For the same zigzag maneuver, \cref{fig:zigzag-maneuver-ood} shows OOD behavior due to an extreme rudder angle of 40°. 
Similarly, in the waypoint navigation case, the AV must navigate through predefined waypoints while following the designated paths. 
Any deviations from these paths, whether due to environmental disturbances or malfunctions, are particularly interesting for testing purposes~\cite{skulstad2021co}. 
\cref{fig:waypoint-following-combined} presents examples of waypoint navigation. 
The 2D path representing IND behavior is depicted in~\cref{fig:waypoint-following-ind}, the 3D path representing IND behavior is illustrated in~\cref{fig:waypoint-following-3d}, and the 2D path demonstrating an OOD occurrence due to sensor noise is shown in~\cref{fig:waypoint-following-ood}. 


To perform zigzag maneuvers, it is necessary to control the rudder systematically.
The vessel’s rudder alternates to either side by an angle of $\delta$ degrees each time the vessel’s heading deviates by $\psi$ degrees from its initial course, as illustrated in~\cref{fig:zigzag-maneuver-ind}. The vessel maneuver begins with the rudder turning by $\delta$ degrees to either the left or right, depending on the initial setup. Once the vessel has shifted $\psi$ degrees away from its starting direction, the rudder is moved to the same angle in the opposite direction. After this change, the vessel initially continues to turn in the original direction but at a decreasing rate, eventually reversing its yaw to follow the new rudder position. When the heading again deviates by $\psi$ degrees from the desired course, the rudder angle switches back, and the cycle repeats. This back-and-forth rudder pattern is defined by the values of $\delta$ (rudder angle) and $\psi$ (heading deviation), commonly represented as $\delta$/$\psi$. 

To design paths for waypoint navigation, we follow a simple generation process that accounts for operational conditions and constraints. 
The generation process begins by specifying several key parameters: the number of waypoints, a radius of acceptance (\textit{R\_switch}) around each waypoint, coordinate ranges for each axis (\textit{x\_range}, \textit{y\_range}, and \textit{z\_range} for 3D paths), an optional minimum distance between consecutive waypoints, and the total number of iterations or sets of waypoints to be generated. The setting of these parameters is inspired by the examples found in the MSS. The only difference is in introducing the optional \textit{min\_distance}, which was necessary in some vessel models where the navigation path was not completed regardless of the \textit{R\_switch}.

\subsubsection{RQ1 and RQ2 Setup}\label{ssec:rq1setup}
Using the specified maneuvers (\cref{ssec:maneuvers}), i.e., zigzag maneuver and waypoint navigation, we created normal operational scenarios for the AVs to collect the IND dataset for training. 
For the same maneuvers, we introduced sensor or rudder noise to generate testing datasets that include both IND and OOD instances. 
Given that each vessel supports distinct controls, DoF, and maneuvers (\cref{tab:vessel_characteristics}), we created paths to analyze the effects of sensor and rudder noise individually.

We generated 30 waypoint navigation paths for each vessel---Mariner, Remus 100, NPS AUV, and Otter. 
Out of these, 20 paths, which represent normal AV operation, were designated to generate training datasets. 
The remaining 10 paths were utilized to generate testing datasets, with introduced sensor noise.
Sensor noise is typically modeled using Gaussian distribution, which reflects real-world characteristics of sensor inaccuracies \cite{lee2024robust}.
Noise is often introduced by adding random values drawn from a Gaussian distribution 
$N(\mu, \sigma^2)$, where $\mu$ represents the mean (commonly set to 0 for unbiased noise) and 
$\sigma^2$ denotes the variance, controlling the noise magnitude.
This approach allows for realistic simulation of sensor errors, enabling systems to be evaluated under varying noise conditions
In our settings, we introduced noise magnitude values from the predefined set 
$[2, 3, 4, 5, 6, 7, 8]$, corresponding to the standard deviation of the Gaussian noise. 
These noise values were incorporated into the $x$ and $y$ positions for 2D paths, and the $x$, $y$, 
and $z$ positions for 3D paths. For each time step, the noisy positions were calculated using \cref{eq-sensor-noise}, defined as:
\begin{equation}\label{eq-sensor-noise}
x_{\text{pos}} = x_{\text{true}} + \sigma \cdot \text{randn}
\end{equation}

where \texttt{randn} generates samples from a standard Gaussian distribution. By varying the noise magnitude 
($\sigma$), we simulated different levels of sensor inaccuracies, ranging from mild to severe.
During the vessel's operation, noise was introduced at a randomly selected point and applied for a 2-minute duration, simulating temporary sensor inaccuracies.
Regarding actuator noise, we employed zigzag maneuvers on vessels already capable of such operations, including Mariner, Container, and Remus 100.
We used rudder angle values within the normal range [10, 15, 20, 30] to generate training datasets. 
Note that this normal rudder angle range is commonly used for vessels' testing with zigzag maneuvers~\cite{wang2021incorporating, kanazawa2022knowledge, abdel2013simulation}. 
To introduce actuator noise, we added rudder angle values from the range [40, 45, 50] to the previously defined normal rudder angles. 
These extreme rudder angle values are introduced for a 2-minute duration.








\subsubsection{RQ3 Setup}\label{ssec:rq2setup}
In RQ3, we aim to analyze OOD behavior due to environmental disturbances. 
As shown in \cref{tab:vessel_characteristics}, only three vessels, i.e., Remus 100, Otter, and NPS AUV, support the ocean current as an environmental disturbance for the waypoint navigation. 
We generated 30 waypoint navigation paths for these vessels. 
Out of these, 20 paths, which represent normal AV operation with default ocean current ($Vc=0.5$), were designated to generate training datasets. 
The remaining 10 paths were utilized to generate testing datasets, with increased ocean current speed ($Vc=0.65$).
We introduced this high ocean current at a randomly chosen point and for a 2-minute duration.



\subsubsection{RQ4 Setup}
To compare our approach, we opted for two traditional and widely used OOD detection methods~\cite{yang2021generalized, gwon2023out, lee2018simple, graham2023latent}, i.e., the Root Mean Squared Error (RMSE) as an output-based method, and the Euclidean distance as a distance based-method. 
We set up these methods as alternatives to DTC of \approach{}. 
Specifically, we devised two configurations: (i) \emph{DTM-R}, which utilizes the DTM of our \approach{} integrated with RMSE, and (ii) \emph{DTM-E}, where we employ the DTM of \approach{} combined with the Euclidean distance.
To calculate the RMSE for \emph{DTM-R}, we used \cref{eq-rmse}, where $\hat{y}_i$ represents a predicted state, $y_i$ denotes the true state, and $n$ is the total number of predictions.
For computing the Euclidean distance in the case of \emph{DTM-E}, we used \cref{eq-eudist}, where $\hat{x_i}$ represents a predicted state, $y_i$ denotes the true state, and $n$ again indicates the total number of predictions. 

\begin{equation}\label{eq-rmse}
RMSE = \sqrt{\frac{1}{n} \sum_{i=1}^{n} \left(y_i - \hat{y}_i \right)^2}
\end{equation}

\begin{equation}\label{eq-eudist}
d_E(x, y) = \sqrt{\sum_{i=1}^{n} (x_i - y_i)^2}
\end{equation}

We utilized training and testing datasets specifically generated for RQ1 and RQ2, following the setups outlined in \cref{ssec:rq1setup} and \cref{ssec:rq2setup}, respectively.
During training, we used \cref{eq-rmse} and \cref{eq-eudist} to calculate the thresholds corresponding to \emph{DTM-R} and \emph{DTM-E}, respectively. 
Since we train models using IND data, the thresholds calculated during training represent IND. 
For testing, we utilized data containing both IND and OOD instances to analyze each method's performance in predicting OOD.  
Specifically, we used all methods to analyze the OOD of each testing data point by comparing the method's output (e.g., reconstruction error) to the thresholds calculated for each method (as described in~\cref{ssec:dtc}).

\subsection{Experiment Execution}

We conducted the experiments on a workstation equipped with an Apple M1 Pro chip, featuring a 10-core CPU (8 performance cores and 2 efficiency cores) and 32 GB of unified memory. 
Model training was performed on the CPU, as the computational requirements were manageable without a dedicated GPU.


\subsection{Evaluation Metrics and Statistical Tests}
\paragraph{Evaluation Metrics} To analyze results for RQ1 and RQ2, we employ two standard out-of-distribution detection metrics: the \emph{Area Under the Receiver Operating Characteristic} (AUROC, \cref{eq:auroc}) and the \emph{True Negative Rate at 95\% True Positive Rate} (TNR@TPR95, \cref{eq:tnr_at_tpr95}),  which are standard in OOD detection studies~\cite{che2021deep, hsu2020generalized}. For RQ3, we compare all approaches (i.e., \approach{}, DTM-R, and DTM-E) using statistical tests appropriate for dichotomous experimental data.

\paragraph{AUROC}
AUROC measures how well a model distinguishes between in-distribution (IND) and out-of-distribution (OOD) samples across all possible decision thresholds. Let \(\text{TPR}(\alpha)\) and \(\text{FPR}(\alpha)\) denote the true positive rate and false positive rate, respectively, at threshold \(\alpha\). Then, the AUROC is defined as:
\begin{equation}
  \label{eq:auroc}
  \text{AUROC} \;=\; \int_{0}^{1} \text{TPR}(\alpha)\, d\bigl(\text{FPR}(\alpha)\bigr).
\end{equation}
This integral represents the area under the ROC curve, capturing the trade-off between \(\text{TPR}(\alpha)\) and \(\text{FPR}(\alpha)\) over all possible thresholds. AUROC can also be interpreted as the probability that a randomly chosen OOD sample receives a higher detection score than a randomly chosen IND sample, making it a key threshold-independent metric for OOD detection.

\paragraph{TNR@TPR95}
TNR@TPR95 targets a specific operating point of high true positive rate (TPR). We first identify the threshold \(\alpha^*\) for which the TPR is closest to \(0.95\):
\begin{equation}
  \alpha^* \;=\; \arg\min_{\alpha} \,\bigl|\text{TPR}(\alpha) - 0.95\bigr|.
\end{equation}
Let \(\text{FPR}(\alpha^*)\) be the false positive rate at that threshold. The \emph{true negative rate} (TNR) is \( 1 - \text{FPR}\), hence:
\begin{equation}
  \label{eq:tnr_at_tpr95}
  \text{TNR@TPR95} \;=\; 1 \;-\; \text{FPR}\bigl(\alpha^*\bigr).
\end{equation}
TNR@TPR95 measures how well the model can maintain a 95\% True Positive Rate for OOD detection (i.e., correctly identifying OOD data) while still preserving a high True Negative Rate for in-distribution samples (i.e., correctly identifying ID data).


\paragraph{Statistical Analyses}
We use hypothesis testing and effect size measures to statistically evaluate our findings. A set of observations is defined as a distribution of metric values, e.g., performance scores across noise magnitudes. Depending on the research question, different statistical tests are applied to assess significant differences and quantify effect sizes.  We follow the best practice~\cite{arcuri2011practical}.
For RQ1 and RQ2, we use the Kruskal-Wallis H test~\cite{kruskal1952use} to determine whether OOD detection performance significantly varies across different noise magnitudes and/or across vessel types. If a significant difference is detected (\(\alpha = 0.05\)), we conduct Dunn’s post-hoc test~\cite{dunn1964multiple} with Bonferroni correction~\cite{dunn1961multiple} to identify which specific pairs of noise levels or vessels exhibit significant differences. 
Additionally, we perform the Vargha-Delaney \vda{} test \cite{vargha2000critique} to assess the effect size magnitude across all comparisons. Two observation sets are considered stochastically equivalent if \( \hat{A}_{12} = 0.5 \). If \( \hat{A}_{12} > 0.5 \), the first observation is stochastically better, whereas \vda{} \(< 0.5 \) indicates the second observation is better. The effect size is further categorized using \( \hat{A}_{12}^{scaled} = (\hat{A}_{12} - 0.5) \times 2 \) as follows~\cite{hess2004robust}:  
\textbf{Negligible}: \( | \hat{A}_{12}^{scaled} | < 0.147 \),
\textbf{Small}: \( 0.147 \leq | \hat{A}_{12}^{scaled} | < 0.33 \),
\textbf{Medium}: \( 0.33 \leq | \hat{A}_{12}^{scaled} | < 0.474 \),
\textbf{Large}: \( | \hat{A}_{12}^{scaled} | \geq 0.474 \).  
Statistical significance is determined when the p-value is below  \(\alpha\) and the effect size is non-negligible. 

To investigate whether performance is consistently monotonic with respect to noise magnitudes, we measure the Spearman rank correlation, a widely used metric that is robust to outliers and does not assume linearity \cite{spearman_correlation}. 
The Spearman correlation test is particularly suitable in this context, as our dataset meets its core assumptions: ordinal noise magnitudes, continuous performance measures, independent observations, and an expected monotonic relationship.
Lastly, for RQ4, we use the Chi-square test with \(\alpha = 0.05\) to assess the statistical significance of performance differences among \approach{}, DTM-R, and DTM-E. We also report Cohen’s \(h\) to quantify the magnitude of these differences.


\subsection{Results}
\subsubsection{RQ1 Results (Effectiveness - Sensor Noise)}

The results in~\cref{tab:rq1_sensor_noise_with_correlation} demonstrate that \approach{} effectively detects out-of-distribution occurrences induced by sensor noise across different AV models. With AUROC values consistently above 95\% across all noise magnitudes and AV types---including Mariner, Remus 100, NPS AUV, and Otter---shows a strong ability of \approach{} to distinguish between normal and noise-affected behaviors. Notably, its performance improves as noise magnitude increases, with AUROC values reaching nearly 99\% for most models at higher noise levels. This trend suggests that larger noise magnitudes create more noticeable deviations from expected behavior, which \approach{} effectively detects as OOD occurrences.

\begin{table}[h!]
\ra{1.2}
\centering
\caption{OOD detection performance across different noise magnitudes (m2–m8) for each vessel, reported in terms of AUROC and TNR@TPR95. Spearman correlation coefficients ($r_s$) are used to assess the relationship between noise magnitude and detection performance, with statistical significance ($p < 0.05$) marked by (\(\star\))}
\resizebox{\textwidth}{!}{%
\begin{tabular}{llccccccc ccl}
\toprule
\multirow{2}{*}{Metric} & \multirow{2}{*}{Vessel} & \multicolumn{7}{c}{Noise Magnitude} & \multicolumn{3}{c}{Correlation} \\
\cmidrule(lr){3-9} \cmidrule(l){10-12}
& & m2 & m3 & m4 & m5 & m6 & m7 & m8 & $r_s$ & p-value & Strength \\
\midrule
\multirow{4}{*}{AUROC} 
& Mariner    & \newPerc{96.03} & \newPerc{94.34} & \newPerc{96.23} & \newPerc{97.19} & \newPerc{97.00} & \newPerc{96.76} & \newPerc{95.79} & 0.02 & $>$ 0.05 & Very weak \\
& Remus 100  & \newPerc{98.58} & \newPerc{99.23} & \newPerc{99.37} & \newPerc{99.37} & \newPerc{99.48} & \newPerc{99.64} & \newPerc{99.53} & 0.33 & $<$ 0.05\textsuperscript{\(\star\)} & Weak \\
& NPS AUV    & \newPerc{97.47} & \newPerc{98.35} & \newPerc{98.72} & \newPerc{98.98} & \newPerc{99.18} & \newPerc{99.34} & \newPerc{99.48} & 0.68 & $<$ 0.05\textsuperscript{\(\star\)} & Strong \\
& Otter      & \newPerc{97.98} & \newPerc{98.52} & \newPerc{98.94} & \newPerc{98.98} & \newPerc{99.06} & \newPerc{99.23} & \newPerc{99.33} & 0.61 & $<$ 0.05\textsuperscript{\(\star\)} & Strong \\
\midrule
\multirow{4}{*}{TNR@TPR95} 
& Mariner    & \newPerc{87.91} & \newPerc{83.36} & \newPerc{89.96} & \newPerc{90.13} & \newPerc{91.96} & \newPerc{91.97} & \newPerc{86.76} & 0.01 & $>$ 0.05 & Very weak \\
& Remus 100  & \newPerc{96.82} & \newPerc{98.27} & \newPerc{98.71} & \newPerc{98.64} & \newPerc{98.78} & \newPerc{99.07} & \newPerc{99.26} & 0.32 & $<$ 0.05\textsuperscript{\(\star\)} & Weak \\
& NPS AUV    & \newPerc{93.80} & \newPerc{95.99} & \newPerc{97.32} & \newPerc{97.71} & \newPerc{98.26} & \newPerc{98.20} & \newPerc{98.56} & 0.70 & $<$ 0.05\textsuperscript{\(\star\)} & Strong \\
& Otter      & \newPerc{95.80} & \newPerc{96.81} & \newPerc{97.39} & \newPerc{97.43} & \newPerc{97.71} & \newPerc{98.01} & \newPerc{98.08} & 0.60 & $<$ 0.05\textsuperscript{\(\star\)} & Strong \\
\bottomrule
\label{tab:rq1_sensor_noise_with_correlation}
\end{tabular}}
\end{table}

However, while AUROC remains stable across noise levels, the TNR@TPR95 metric, which reflects the model’s specificity (correctly classifying true negatives) at high sensitivity (correctly classifying true positives), shows some variability, especially at lower noise levels. For example, Mariner’s TNR@TPR95 starts around 87\% at m2, indicating that subtle noise may pose challenges for maintaining a high balance between true negatives (correctly classified in-distribution samples) and true positives (correctly classified OOD samples). This variability in TNR@TPR95 at lower noise levels may come from two main factors. First, when noise is minimal, the small differences make it harder for the model to tell OOD events apart from normal variations, causing overlap between regular and abnormal data. Second, when the noise happens during complex moments, like switching waypoints, it can look similar to normal control changes (e.g., rudder movements), making it challenging for the model to separate expected adjustments from actual OOD events.
Nevertheless, as noise magnitude increases, TNR@TPR95 stabilizes and improves across all models, reinforcing \approach{}’s robustness under more severe noise conditions. Furthermore, performance varies slightly across AV types, with Remus 100 achieving particularly high scores, suggesting that \approach{} is adaptable and consistently reliable across different architectures. Overall, these results underscore \approach{}’s capacity to detect OOD events caused by sensor noise, effectively flagging potential disruptions in AV operation. Below, we analyze the
results based on the three key aspects defined in our research question:

\begin{figure}[tbp]
\scriptsize
\vspace{0.2em}
  \centering
    \includegraphics[width=1\columnwidth]{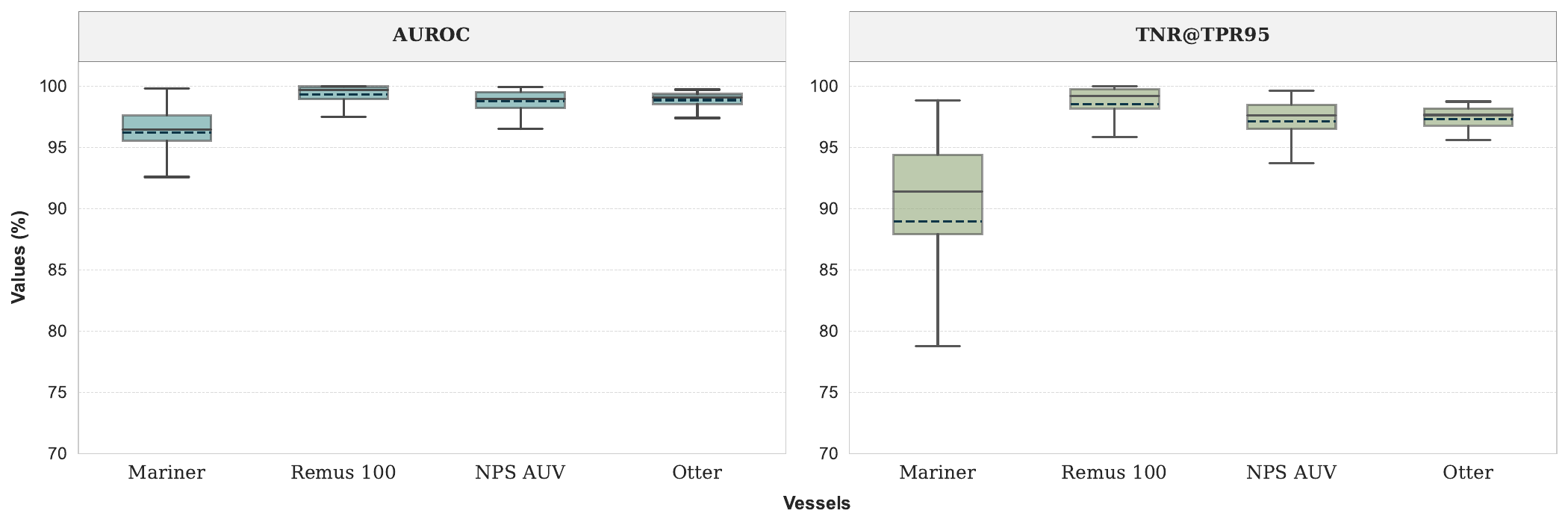}
    \caption{Vessel-wise \textit{AUROC} and \textit{TNR@TPR95} performance for OOD detection for all noise magnitudes combined across respective AVs. The y-axis displays scores in percentage, while the x-axis represents different vessel models we compare.}
    \label{fig:auroc-boxplot-ship-wise}
\end{figure}

\begin{figure}[tbp]
\vspace{0.2em}
  \centering
    \includegraphics[width=1\columnwidth]{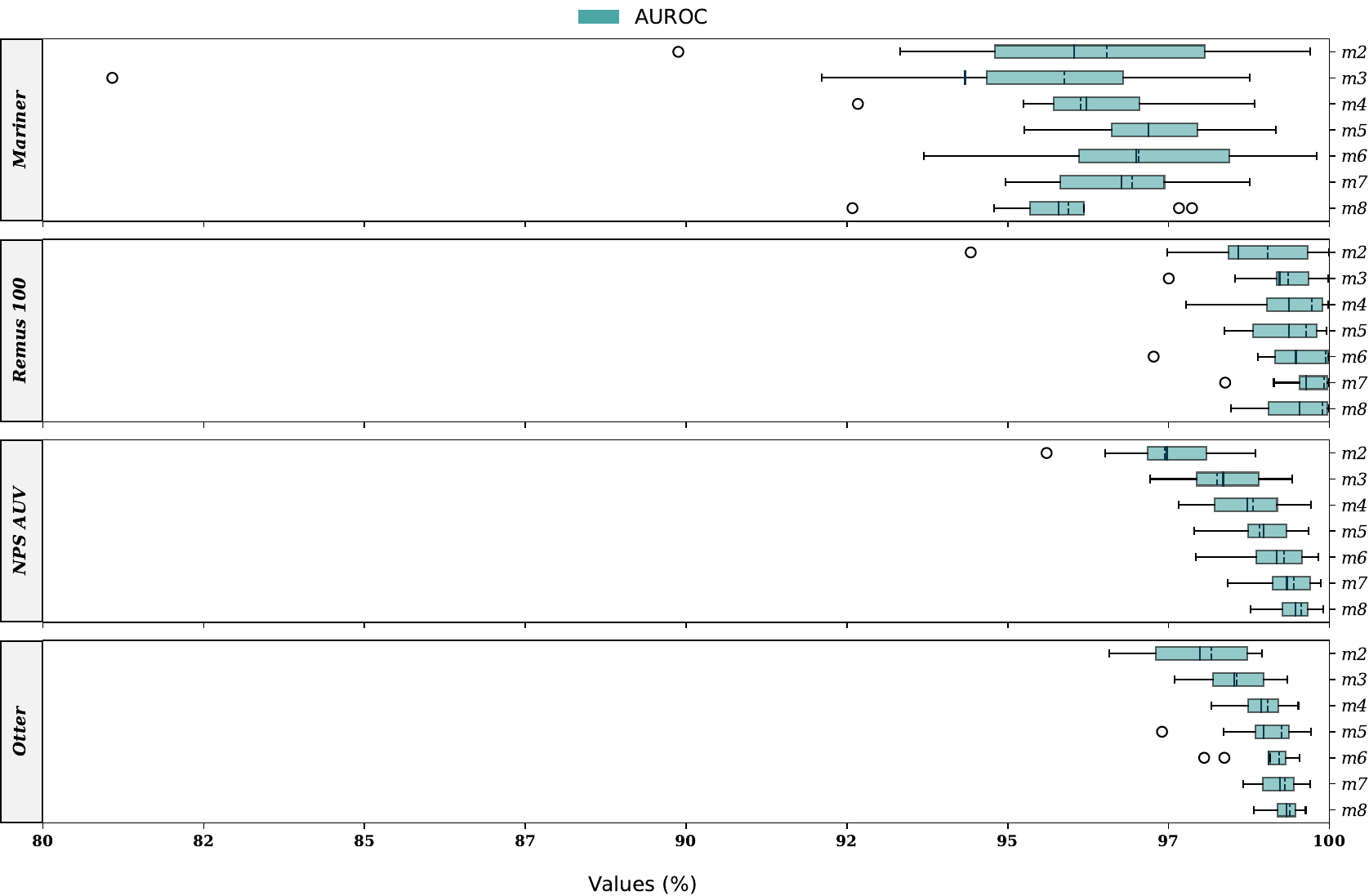}
    \caption{Vessel-wise \textit{AUROC} performance for OOD detection across different noise magnitudes, grouped by ship models. The y-axis represents the noise magnitude levels (m2–m8), while the x-axis displays scores in percentage.}
    \label{fig:ship-wise-auroc-boxplot}
\end{figure}

\begin{figure}[tbp]
\vspace{0.2em}
  \centering
    \includegraphics[width=1\columnwidth]{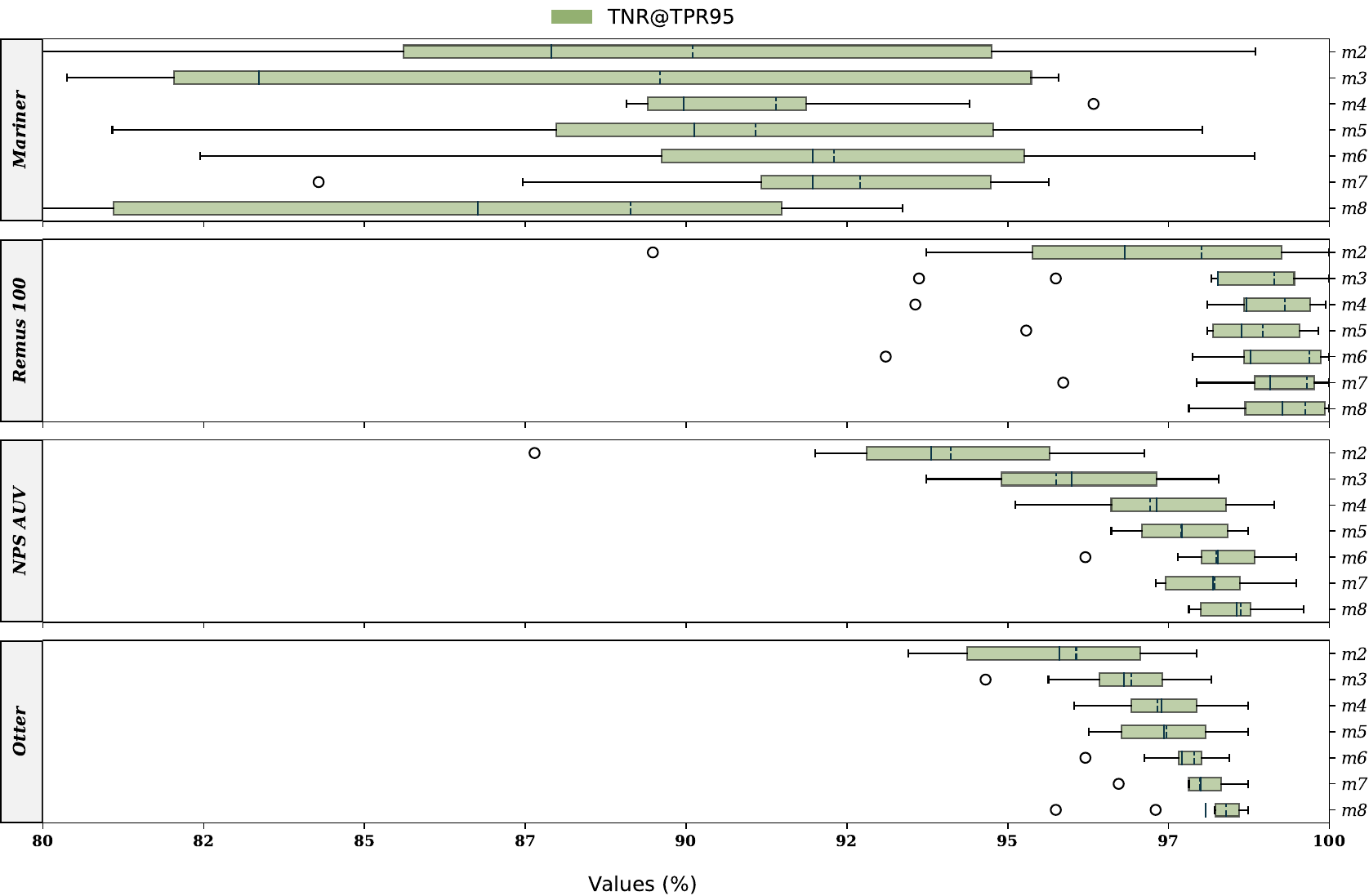}
    \caption{Vessel-wise \textit{TNR@TPR95} performance for OOD detection across different noise magnitudes, grouped by ship models. The y-axis represents the noise magnitude levels (m2–m8), while the x-axis displays scores in percentage.}
    \label{fig:ship-wise-tnr-boxplot}
\end{figure}

\paragraph{\textbf{a) Vessel Types.}} 
\Cref{fig:auroc-boxplot-ship-wise} presents the AUROC and TNR@TPR95 scores for different vessels (Mariner, Remus 100, NPS AUV, and Otter), summarizing overall performance without distinguishing between individual noise magnitudes. The boxplots illustrate the central tendency, variability, and consistency of \approach{}’s performance across vessel types.

The AUROC results indicate that Remus 100, NPS AUV, and Otter exhibit consistently high performance with minimal variation. This suggests that \approach{} is highly effective at distinguishing OOD instances for these vessels, maintaining stable detection regardless of sensor-related uncertainties. Mariner, however, shows greater performance variability, with a wider spread in AUROC scores, indicating that \approach{}'s ability to separate OOD from IND data is less consistent for this vessel.
The TNR@TPR95 results further reinforce this trend. While Remus 100, NPS AUV, and Otter maintain high and stable true negative rates, Mariner exhibits significantly higher variability. This suggests that \approach{} struggles more with false positives for Mariner, leading to less reliable OOD detection performance in this case.

These observations are further supported by the statistical tests in \cref{tab:rq1a_oddit_statistical_results}, where Mariner is consistently outperformed by the other vessels in both AUROC and TNR@TPR95. Additionally, the statistical comparisons show no significant difference between NPS AUV and Otter, reinforcing the findings from the boxplots.

Based on these results, the vessel ranking (ordering from best to worst) in terms of ODDIT’s performance is as follows:
\begin{inparaenum}[(1)]
    \item Remus 100,
    \item NPS AUV and Otter, and
    \item Mariner.
\end{inparaenum}
The same ranking applies to both AUROC and TNR@TPR95 metrics.


\begin{tcolorbox}[colframe=black!50, colback=gray!5, boxrule=0.3mm]
\approach{} performs consistently well for Remus 100, NPS AUV, and Otter, with stable AUROC and TNR@TPR95 scores, indicating reliable OOD detection. Mariner, however, shows high variability, struggling more with false positives. Statistical tests confirm these trends, ranking vessels from best to worst as (1) Remus 100, (2) NPS AUV and Otter, and (3) Mariner, for both metrics.
\end{tcolorbox}

\begin{figure}[tbp]
\scriptsize
\vspace{0.2em}
  \centering
    \includegraphics[width=1\columnwidth]{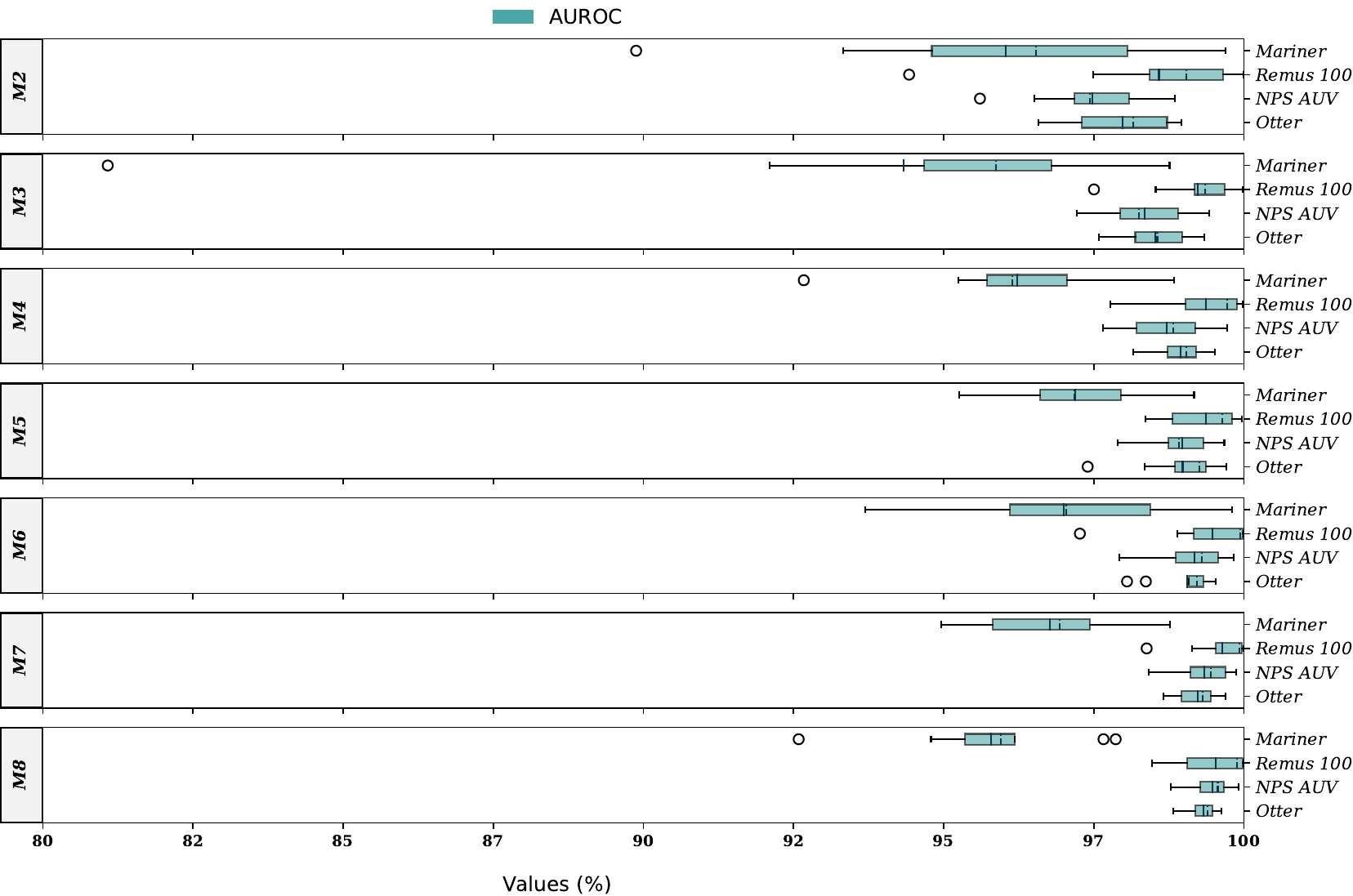}
    \caption{Noise-wise AUROC performance for OOD detection across different ship models, grouped by noise magnitudes. The y-axis represents the ship models (Mariner, Remus 100, NPS AUV, and Otter), while the x-axis displays scores in percentages.}
    \label{fig:auroc-boxplot-noise-wise}
\end{figure}

\begin{figure}[htb]
\centering
\scriptsize
\vspace{0.2em}
    \includegraphics[width=1\columnwidth]{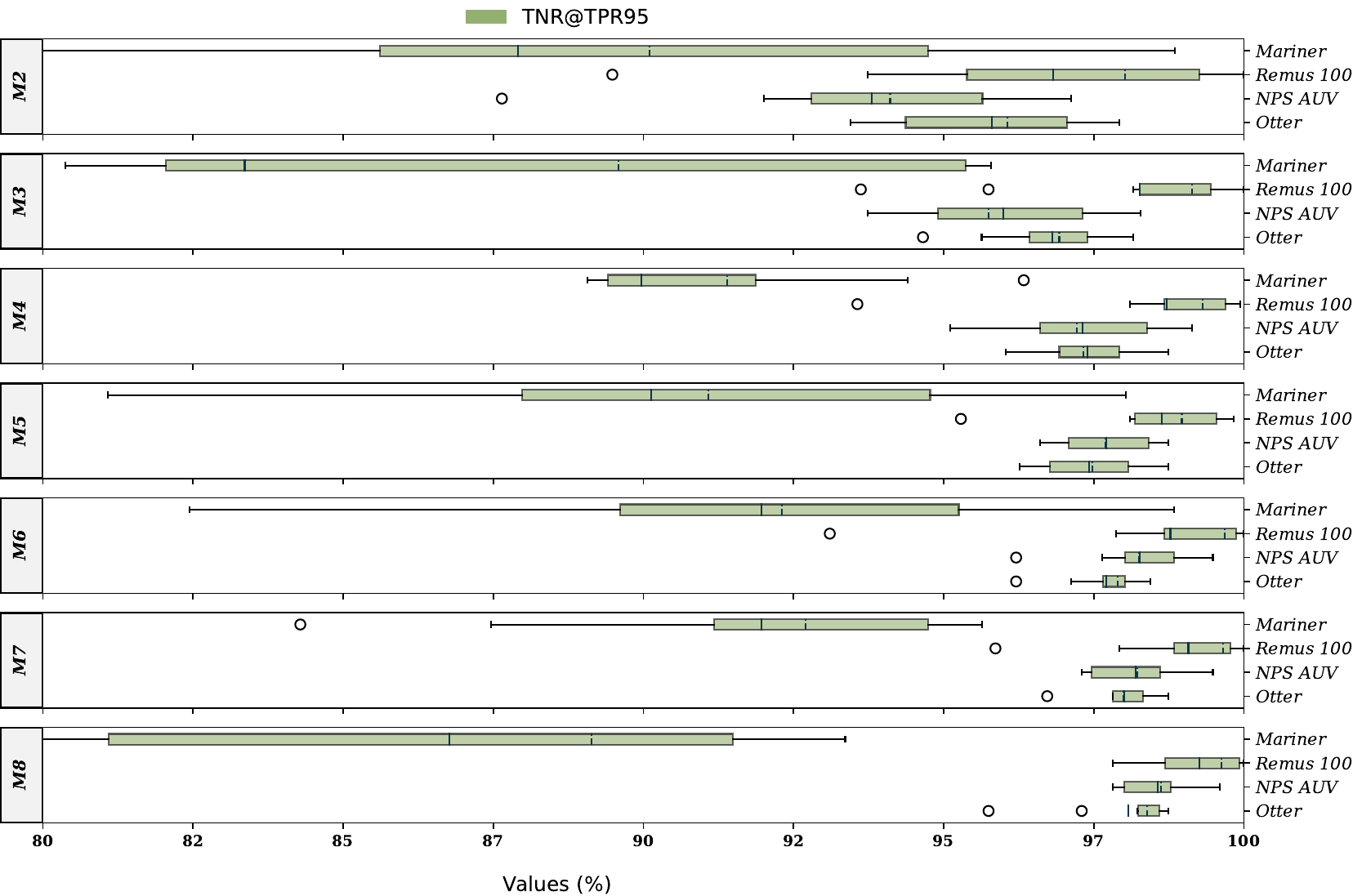}
    \caption{Noise-wise TNR@TPR95 performance for OOD detection across different ship models, grouped by noise magnitudes. The y-axis represents the ship models (Mariner, Remus 100, NPS AUV, and Otter), while the x-axis displays scores in percentages.}
    \label{fig:tnr@tpr95-boxplot-noise-wise}
\end{figure}

\paragraph{\textbf{b) Noise Magnitude.}} 

To provide a clearer comparison of how OOD detection performance varies with increasing noise levels across different AV models, we present the results in \cref{fig:ship-wise-auroc-boxplot,fig:ship-wise-tnr-boxplot}. The boxplots provide insights into the central tendency, variability, and performance distribution for each vessel under different noise scenarios. The AUROC (\cref{fig:ship-wise-auroc-boxplot}) highlights that all ships maintain consistently high AUROC values across noise levels, reflecting their robustness in distinguishing OOD instances. Notably, higher noise magnitudes (e.g., m6 to m8) lead to marginally improved performance for these ships, as evidenced by narrower interquartile ranges and higher medians. On the other hand, Mariner exhibits higher variability, with performance peaks at specific noise levels (e.g., m5 and m6) but less consistency overall. The TNR@TPR95 (\cref{fig:ship-wise-tnr-boxplot}) reveals similar trends, with Remus 100, NPS AUV, and Otter showing high and stable performance regardless of noise magnitude.

Across vessels, there is a general trend where higher noise magnitudes (e.g., m6 to m8) lead to improved detection performance, particularly for NPS AUV and Otter. This observation is further supported by the statistical tests in \cref{tab:rq1b_noise_magnitude_comparison_npsauv,tab:rq1b_noise_magnitude_comparison_otter}, which confirm that performance increases progressively from lower to higher noise magnitudes for these vessels.
Notably, no significant differences were found between noise magnitudes for Mariner and Remus 100, indicating that our approach performs equally across all noise levels for these vessels for both AUROC and TNR@TPR95. 

Based on these findings, the ranking (ordering from best to worst) of noise magnitude impact on \approach{} performance is as follows:

\begin{itemize}
    \item \textit{NPS AUV Vessel (AUROC)}:
    \begin{inparaenum}[(1)]
        \item m8
        \item m6, m7
        \item m4, m5
        \item m3
        \item m2
    \end{inparaenum}

    \item \textit{NPS AUV Vessel (TNR@TPR95)}:
    \begin{inparaenum}[(1)]
        \item m6, m8
        \item m5, m7
        \item m4
        \item m3
        \item m2
    \end{inparaenum}

    \item \textit{Otter Vessel (same order/ranking applies to both metrics)}:
    \begin{inparaenum}[(1)]
        \item m8
        \item m7
        \item m4, m5, m6
        \item m3
        \item m2
    \end{inparaenum}
\end{itemize}

These rankings indicate that higher noise magnitudes (e.g., m8, m6, and m7) tend to improve \approach{}’s detection performance, while lower noise magnitudes (e.g., m2, m3) result in less effective OOD detection. The ordering remains consistent for the Otter vessel, whereas the NPS AUV exhibits slight variations between AUROC and TNR@TPR95 rankings.

\begin{tcolorbox}[colframe=black!50, colback=gray!5, boxrule=0.3mm]
Higher noise magnitudes (m6 to m8) improve \approach{}’s OOD detection, especially for NPS AUV and Otter, while Mariner shows inconsistent performance. Remus 100 remains unaffected. Lower noise levels (m2, m3) lead to weaker detection.
\end{tcolorbox}

\paragraph{\textbf{c) Vessel and Noise.}}
\Cref{fig:auroc-boxplot-noise-wise,fig:tnr@tpr95-boxplot-noise-wise}
show the performance of \approach{} across different ship models (Mariner, Remus 100, NPS AUV, and Otter) using the AUROC and TNR@TPR95 metrics for noise levels ranging from m2 to m8. 
For AUROC (\cref{fig:auroc-boxplot-noise-wise}), higher noise magnitudes (m6 to m8) consistently lead to better performance for all ships, with NPS AUV and Remus 100 showing median values above 99\% and narrow interquartile ranges, indicating high reliability in detecting OOD instances. Otter similarly achieves strong results but exhibits slightly wider ranges, particularly at m2 and m3. In contrast, Mariner’s performance fluctuates more noticeably, with a broader range and lower median at M3 (around 94\%) and less consistent improvement with noise magnitude. 
For TNR@TPR95 (\cref{fig:tnr@tpr95-boxplot-noise-wise}), the trend of better performance with higher noise is also evident, though Mariner exhibits significant variability across all noise levels, with a median value below 90\% for m2 and broader interquartile ranges across most noise magnitudes. Remus 100 and NPS AUV, in contrast, achieve consistently high TNR@TPR95 values, exceeding 98\% at m6 and above, with minimal variability. Otter performs comparably, with steady improvements from m2 to m8 but slightly wider ranges at lower noise levels. Overall, these results demonstrate that the proposed approach performs reliably across ships, with higher noise magnitudes improving OOD detection performance, although variability is more pronounced for certain ships like Mariner.



\Cref{tab:oddit-vessel-ranking} presents the rankings of different vessels—Remus 100 (R), NPS AUV (N), Otter (O), and Mariner (M)—based on the \approach{} performance across noise magnitudes (m2 to m8). The rankings are determined separately for AUROC and TNR@TPR95, providing insights into how \approach{} detects OOD states across different vessels and noise conditions (more details can be found in \cref{tab:rq1c_statistical_results}).

\begin{tcolorbox}[colframe=black!50, colback=gray!5, boxrule=0.3mm]
\approach{} performs best on Remus 100, maintaining high and stable detection across all noise levels. NPS AUV and Otter show comparable performance, with slight variability at lower noise levels (m2, m3). Mariner has the weakest and most inconsistent detection, with greater variability across noise magnitudes.
\end{tcolorbox}

\begin{table}[h]
    \centering
    \caption{Vessel rankings based on \approach{} performance across noise magnitudes for AUROC and TNR@TPR95. 
    The vessels are denoted as follows: R = Remus 100, N = NPS AUV, O = Otter, and M = Mariner.}
    \label{tab:oddit-vessel-ranking}
    \renewcommand{\arraystretch}{1.2} 
    \begin{tabular}{c ccccccc}
        \toprule
        \multirow{2}{*}{Rank} & \multicolumn{7}{c}{Noise Magnitude} \\
        \cmidrule(lr){2-8}
        & m2 & m3 & m4 & m5 & m6 & m7 & m8 \\
        \midrule
        \multicolumn{8}{c}{AUROC Ranking} \\
        \midrule
        1st & R       & R, O       & R, N, O     & R, N, O     & R             & R, N, O     & R, N, O     \\
        2nd & N, O    & N          & M           & M           & N, O          & M           & M           \\
        3rd & M       & M          &             &             & M             &             &             \\
        \midrule
        \multicolumn{8}{c}{TNR@TPR95 Ranking} \\
        \midrule
        1st & R       & R, O       & R, N, O     & R, N        & R, N        & R, N, O     & R, N, O     \\
        2nd & N, O    & N          & M           & O           & O           & M           & M           \\
        3rd & M       & M          &             & M           & M           &             &             \\
        \bottomrule
    \end{tabular}
\end{table}

\paragraph{\textbf{d) Correlation Analysis.}} 

To further support the observed trends, we analyze the relationship between noise magnitude and OOD detection performance using statistical correlation measures. The results indicate that while higher noise magnitudes generally contribute to improved OOD detection performance, the strength of this relationship varies across different AV models.
The statistical correlation results (Table~\ref{tab:rq1_sensor_noise_with_correlation}) reveal that NPS AUV and Otter exhibit strong positive correlations ($p$-value < 0.05) between noise magnitude and AUROC/TNR@TPR95, indicating that increasing noise levels enhance OOD detection performance for these vessels. This suggests that \approach{} benefits from noise-induced deviations in these cases, making OOD instances more distinguishable.
In contrast, Remus 100 consistently achieves the highest AUROC and TNR@TPR95 scores across all noise levels but shows only a weak correlation with noise magnitude. This suggests that its strong performance is not primarily driven by noise increases but rather by inherent vessel characteristics, such as sensor stability and control dynamics. Similarly, Mariner exhibits very weak correlation and fluctuating performance, indicating that noise alone is not a key determinant of its OOD detection effectiveness.
These findings suggest that while noise magnitude plays a role in improving OOD detection, its impact is vessel-dependent. For some AVs, higher noise levels amplify deviations, making OOD instances easier to detect, while for others, performance remains consistently high regardless of noise.

\begin{tcolorbox}[colframe=black!50, colback=gray!5, boxrule=0.3mm]
\approach{} performs best on Remus 100, regardless of noise. For NPS AUV and Otter, higher noise improves detection due to strong positive correlation. Mariner lacks significant correlation, showing unstable performance.
\end{tcolorbox}


\begin{myexamplec}{\textit{\textbf{RQ1 Summary:}}}
\approach{} effectively detects OOD events caused by sensor malfunctions, achieving high AUROC and TNR@TPR95 across vessels. While increasing noise magnitude generally enhances detection, its impact varies. Strong correlations suggest that higher noise makes OOD instances more distinguishable in some cases, while in others, detection remains high regardless of noise. This indicates that \approach{}'s performance is also influenced by factors such as the operational context and control dynamics of each vessel.
\end{myexamplec}

\subsubsection{RQ2 Results (Effectiveness - Actuator Noise)}

Considering the normal rudder angles (10° to 30°), the results in~\cref{tab:rq1_zigzag}, provide insights into how OOD detection varies when introducing actuator noise at 40°, 45°, and 50°. 
Lower angles represent more typical maneuvering conditions, while higher angles simulate scenarios where the rudder undergoes sharp shifts, creating conditions similar to noise-induced deviations. This range of rudder angles allows us to examine how well the OOD detection model distinguishes between expected, in-distribution behaviors and potential anomalies under more extreme steering conditions. The results show that while all vessels perform well at moderate angles, both Mariner and Container struggle to maintain high specificity (correctly classifying in-distribution as non-OOD) at higher angles, where the natural complexity of sharp turns resembles noise, leading to an increase in false positives.


\begin{table}[h!]
\centering
\scriptsize
\caption{OOD detection results corresponding to actuator noise (i.e., rudder angles)}
\resizebox{\textwidth}{!}{%
\begin{tabular}{lcccccccc}
\toprule
\multirow{2}{*}{Vessel}  & \multicolumn{2}{c}{10°} & \multicolumn{2}{c}{15°} & \multicolumn{2}{c}{20°} & \multicolumn{2}{c}{30°} \\
\cmidrule(lr){2-3} \cmidrule(lr){4-5} \cmidrule(lr){6-7} \cmidrule(lr){8-9}
& AUROC & TNR@TPR95 & AUROC & TNR@TPR95 & AUROC & TNR@TPR95 & AUROC & TNR@TPR95 \\
\midrule
Mariner & \perc{0.9838} & \perc{0.9163} & \perc{0.9972} & \perc{0.9878} & \perc{0.9791} & \perc{0.8842} & \perc{0.9117} & \perc{0.6473} \\
Container & \perc{0.9438} & \perc{0.8424} & \perc{0.9367} & \perc{0.7866} & \perc{0.9555} & \perc{0.8920} & \perc{0.9219} & \perc{0.6844} \\
Remus 100 & \perc{0.9416} & \perc{0.7117} & \perc{0.9758} & \perc{0.8880} & \perc{0.9901} & \perc{0.9816} & \perc{0.9814} & \perc{0.9334} \\
\bottomrule
\label{tab:rq1_zigzag}
\end{tabular}}
\end{table}

The results for Remus 100; however, show an unusual trend, with the lowest TNR@TPR95 at 10° (71.17\%), even though detection should theoretically be easier at this angle due to the clearer noise impact. Although the exact cause remains uncertain, we can consider a few plausible facts that may contribute to this unexpected behavior. One possible explanation lies in the unique dynamics and characteristics of the Remus 100 AUV. Unlike Mariner and Container, which are massive vessels—160m and 175m in length and weighing 17,045 and 21,750 tonnes, respectively—Remus 100 is only 1.6m long and weighs 32 kg. This extreme difference in size and mass means that Remus 100, operating in deep water (differently from Mariner and Container), exhibits much sharper and more frequent turns during zigzag maneuvers. Its smaller size and lighter weight allow for quick directional changes, creating more erratic movement patterns than those of the larger vessels. 
Additionally, the DTM may not have learned Remus 100's patterns at the lowest angle as the training data heavily emphasizes maneuvers between 15° and 30° angle---as demonstrated by the performance improvement at higher angles. 

Below, we analyze the results based on the three key aspects defined in our research question:

\begin{figure}[tbp]
\scriptsize
\vspace{0.2em}
  \centering
    \includegraphics[width=1\columnwidth]{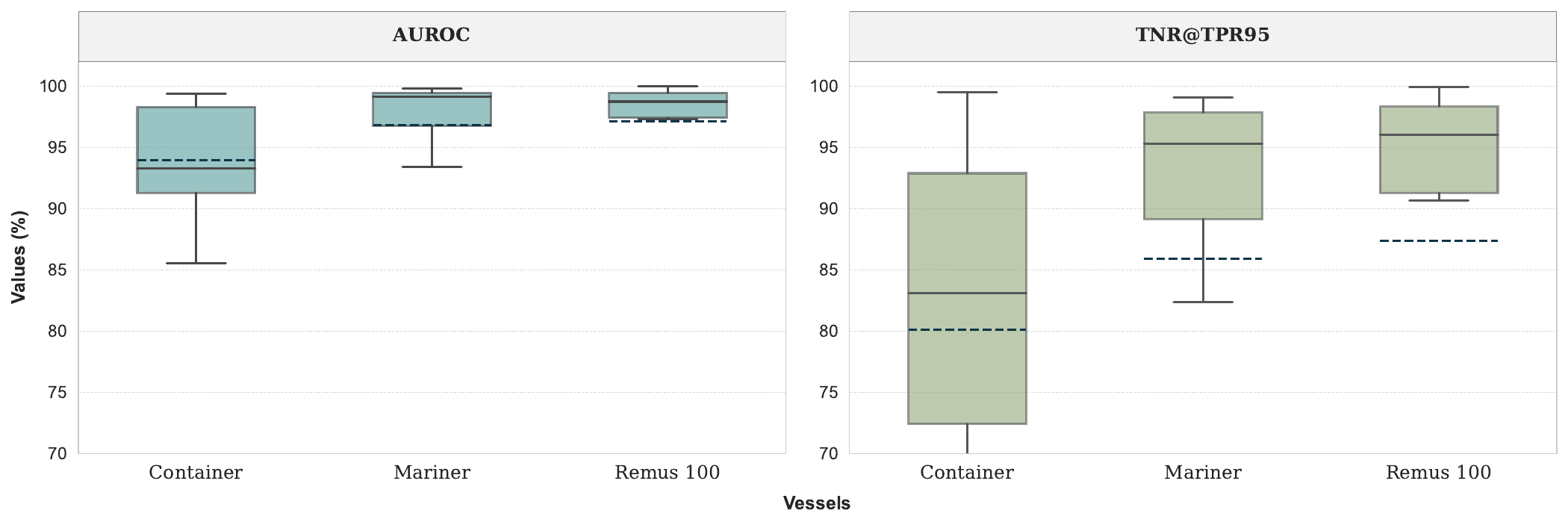}
    \caption{Vessel-wise \textit{AUROC} and \textit{TNR@TPR95} performance for OOD detection for all rudder angles combined across respective AVs. The y-axis displays scores in percentage, while the x-axis represents different vessel models we compare.}
    \label{fig:rq2a-auroc-boxplot-ship-wise}
\end{figure}

\paragraph{\textbf{a) Vessel Types.}} 

\Cref{fig:rq2a-auroc-boxplot-ship-wise} presents the vessel-wise performance of \approach{} across different AVs by aggregating results across all rudder angles. The AUROC boxplot indicates that Mariner and Remus 100 exhibit consistently high performance, with minimal variability and median scores near 100\%. In contrast, the Container vessel shows greater performance variability, with a wider interquartile range and a lower median score, suggesting that \approach{} struggles more with distinguishing OOD instances for this vessel. A similar pattern emerges in the TNR@TPR95 results.

Despite these observed trends, pairwise statistical comparisons using Dunn’s test did not reveal significant differences $(p > 0.05)$ between vessels, indicating that their OOD detection capabilities are statistically comparable. Therefore, establishing a definitive ranking of vessel-wise performance is not possible.

\begin{tcolorbox}[colframe=black!50, colback=gray!5, boxrule=0.3mm]
\approach{} performs consistently well across vessels, with minimal variability for Mariner and Remus 100. Container shows more fluctuations, but statistical tests (p > 0.05) indicate no significant performance differences.
\end{tcolorbox}

\begin{table}[h!]
\centering
\scriptsize
\caption{OOD detection performance across extreme rudder angles (i.e., 40°, 45°, 50°) for each vessel, reported in terms of AUROC and TNR@TPR95.
Spearman correlation coefficients ($r_s$) are used to assess the relationship between rudder angle and detection performance, with statistical significance ($p < 0.05$) marked by (\(\star\))}
\resizebox{0.8\textwidth}{!}{%
\begin{tabular}{llccc ccl}
\toprule
\multirow{2}{*}{Metric} & \multirow{2}{*}{Vessel} & \multicolumn{3}{c}{Rudder Angle} & \multicolumn{3}{c}{Correlation} \\ 
\cmidrule(lr){3-5} \cmidrule(lr){6-8}
& & 40° & 45° & 50° & $r_s$ & p-value & Strength \\ 
\midrule
\multirow{3}{*}{AUROC} 
    & Mariner   & 97.92\%  & 97.79\%  & 94.67\%  & 0.09  & $>$ 0.05 & Very weak\textsuperscript{\(\neg\)} \\
    & Container & 93.32\%  & 91.88\%  & 96.64\%  & 0.27  & $>$ 0.05 & Weak \\
    & Remus 100 & 99.23\%  & 98.56\%  & 92.45\%  & 0.65  & $<$ 0.05\textsuperscript{\(\star\)} & Strong\textsuperscript{\(\neg\)} \\
\midrule
\multirow{3}{*}{TNR@TPR95} 
    & Mariner   & 86.92\%  & 95.28\%  & 75.46\%  & 0.27  & $>$ 0.05 & Weak \\
    & Container & 76.63\%  & 74.60\%  & 89.18\%  & 0.33  & $>$ 0.05 & Weak  \\
    & Remus 100 & 96.96\%  & 94.82\%  & 64.65\%  & 0.58  & $>$ 0.05 & Moderate\textsuperscript{\(\neg\)} \\
\bottomrule
\label{tab:rq1_zigzag_correlation}
\end{tabular}%
}
\end{table}

\paragraph{\textbf{b) Noise Magnitude.}} 


The expectation that higher noise magnitudes (rudder angles) improve OOD detection is not consistently observed. Instead, the results show that performance can decline at extreme rudder angles, particularly for Remus 100 and Mariner. 
As shown in \cref{tab:rq1_zigzag_correlation}, both AUROC and TNR@TPR95 decrease at the highest rudder angle (50°) for these vessels, contradicting the assumption that increased corruption enhances detection.

For Remus 100, the significant drop in AUROC and TNR@TPR95 at 50° suggests that extreme rudder angles introduce instability rather than reinforcing OOD separability. Similarly, for Mariner, the decline in TNR@TPR95 at 50° suggests that higher noise magnitudes compromise the \approach{}’s specificity, causing it to misclassify normal states as OOD more frequently.
On the other hand, the Container model shows an opposite trend, with TNR@TPR95 increasing at 50°, indicating that for this vessel, higher noise magnitudes may contribute positively to specificity. These findings demonstrate that the impact of noise magnitude on OOD detection is not uniform across vessels, highlighting the need for vessel-specific tuning of detection thresholds to account for noise variability.

However, despite these observed variations, statistical tests do not confirm a significant effect of noise magnitude on performance. The Kruskal-Wallis test returned non-significant results $(p > 0.05)$ across all observations, indicating that the observed trends are not strong enough to conclude systematic performance changes across noise levels. 
Therefore, a definitive ranking of rudder angle effects on OOD detection cannot be established.

\begin{tcolorbox}[colframe=black!50, colback=gray!5, boxrule=0.3mm]
Higher noise magnitudes do not consistently improve OOD detection. Extreme rudder angles (50°) reduce AUROC and TNR@TPR95 for Remus 100 and Mariner, introducing instability and misclassifications. In contrast, Container benefits from increased noise. However, statistical tests (p > 0.05) show no significant overall effect, preventing a definitive ranking of rudder angle impact.
\end{tcolorbox}

\paragraph{\textbf{c) Vessel and Noise.}}
\Cref{tab:rq1_zigzag_correlation} presents the OOD detection performance across three vessels (Mariner, Container, and Remus 100) at different extreme rudder angles (40°, 45°, and 50°). We observe that the OOD detection performance of \approach{} varies across vessels at extreme rudder angles. Remus 100 achieves the highest AUROC at 40° but experiences a significant drop in both AUROC and TNR@TPR95 at 50°, indicating that extreme noise levels introduce instability. Mariner maintains relatively stable AUROC but suffers a notable decline in TNR@TPR95 at 50°, suggesting increased false positives. Container shows the weakest AUROC performance overall but demonstrates improved TNR@TPR95 at 50°, suggesting better specificity at extreme rudder angles.
This suggests that vessel characteristics play a key role in OOD detection, with some models being more affected by actuator noise than others.

Focusing on the statistical analysis, we observe that the only significant difference $(p < 0.05)$ occurs at 40° rudder angle, where Remus 100 outperforms Container with a large effect size in both AUROC and TNR@TPR95 (see~\cref{tab:rq2c_ood_comparison}). No other pairwise comparisons reveal significant differences across vessels, indicating that, apart from this case, the performance differences between vessels are not statistically strong. Based on this observation, we establish the following ranking at 40° rudder angle:
\begin{inparaenum}[(1)]
    \item Remus 100,
    \item Mariner, and
    \item Container.
\end{inparaenum}

\begin{tcolorbox}[colframe=black!50, colback=gray!5, boxrule=0.3mm]
\approach{}’s performance varies across vessels at extreme rudder angles. Remus 100 performs best at 40° but drops at 50°, while Mariner remains stable in AUROC but sees more false positives. Container has the weakest AUROC but improves TNR@TPR95 at 50°. Statistical tests (p < 0.05) show a significant difference only at 40°, where Remus 100 outperforms Container.
\end{tcolorbox}

\paragraph{\textbf{d) Correlation Analysis.}} 
The correlation analysis reveals that noise magnitude does not have a consistent relationship with OOD detection performance. Remus 100 exhibits the strongest negative correlation between rudder angle and AUROC (-0.65, p < 0.05) and TNR@TPR95 (-0.58, p > 0.05), confirming that higher rudder angles degrade performance rather than improve it. This suggests that at extreme noise levels, actuator noise disrupts separability rather than enhancing it.

In contrast, Mariner and Container ships both show weak and non-significant correlations, indicating that noise magnitude does not strongly impact their detection performance trends. The weak Spearman correlations (-0.09 for AUROC and 0.27 for TNR@TPR95 for Mariner; 0.27 for AUROC and 0.33 for TNR@TPR95 for Container) suggest that for these vessels, performance fluctuations at different rudder angles are not necessarily correlated with noise magnitude.

The lack of strong positive correlations across vessels further supports the finding that extreme rudder angles do not improve OOD detection. Instead, the impact of actuator noise is vessel-dependent, reinforcing the need for tailored detection strategies that account for vessel-specific noise effects.

\begin{tcolorbox}[colframe=black!50, colback=gray!5, boxrule=0.3mm]
Noise magnitude has no consistent impact on OOD detection. Remus 100 shows a significant negative correlation, while Mariner and Container exhibit weak, non-significant trends. Results highlight vessel-dependent noise effects.
\end{tcolorbox}


\begin{myexamplec}{\textit{\textbf{RQ2 Summary:}}}
\approach{} demonstrates strong OOD detection performance under actuator noise, but its effectiveness varies across vessels and rudder angles. The observed variability can be attributed to three key factors: \begin{inparaenum}[(1)] \item \textit{vessel characteristics} (size and maneuverability affecting detection stability), \item \textit{noise intensity} (higher angles degrade performance for some vessels), and \item \textit{model training data} (imbalanced representation at certain angles impacting performance consistency). \end{inparaenum}
\end{myexamplec}




\subsubsection{RQ3 Results (Environmental Disturbances)}

Based on the results in \cref{tab:rq2_ocean_current}, \approach{} demonstrates high accuracy in detecting OOD occurrences caused by environmental disturbances across all tested vessels, i.e., Remus 100, NPS AUV, and Otter. The proposed DT-based approach achieves near-perfect AUROC scores, indicating strong capability in distinguishing in-distribution from out-of-distribution data when faced with the ocean current. Similarly, for TNR@TPR95, \approach{} can maintain high specificity, effectively avoiding false positives while detecting OOD instances reliably.

\begin{table}[h!]
\centering
\scriptsize
\caption{OOD detection results based on environmental disturbances (i.e., ocean current)}
\resizebox{0.7\textwidth}{!}{%

\begin{tabular}{ccc ccc ccc}
\toprule
\multicolumn{2}{c}{\textbf{Remus 100}} & & \multicolumn{2}{c}{\textbf{NPS AUV}} & & \multicolumn{2}{c}{\textbf{Otter}} \\
\cmidrule(lr){1-2} \cmidrule(lr){4-5} \cmidrule(lr){7-8}
\textbf{AUROC} & \textbf{TNR@TPR95} & & \textbf{AUROC} & \textbf{TNR@TPR95} & & \textbf{AUROC} & \textbf{TNR@TPR95} \\
\midrule
\perc{0.9991} & \perc{0.9972} & & \perc{0.9990} & \perc{0.9999} & & \perc{0.9998} & \perc{0.9995} \\
\bottomrule
\label{tab:rq2_ocean_current}

\end{tabular}}
\end{table}

\begin{figure}[h]
\centering
\begin{subfigure}[b]{0.45\columnwidth}
  \centering
  \includegraphics[width=\textwidth]{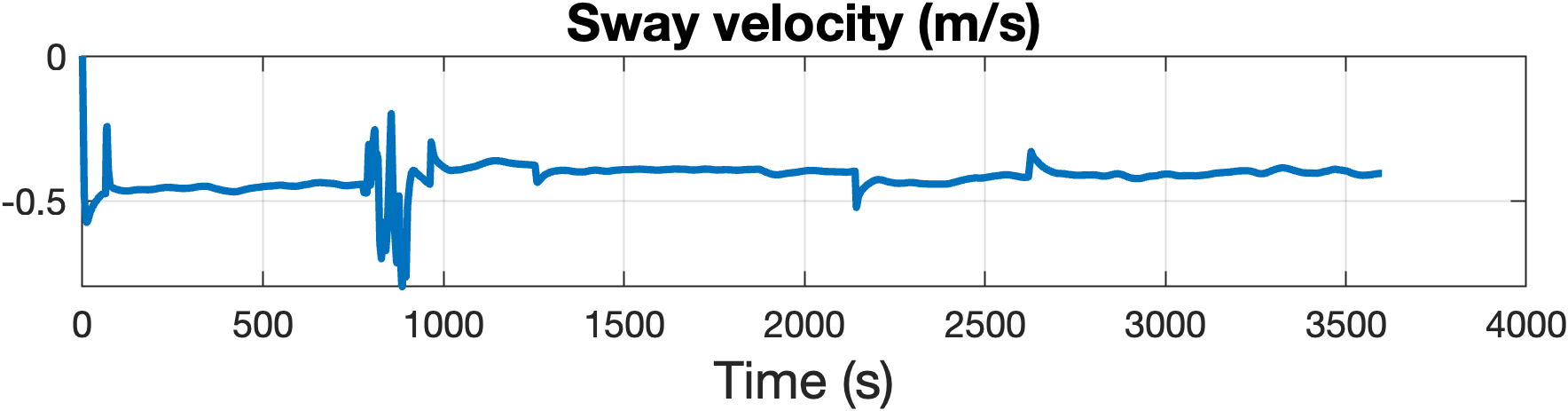}
  \caption{Sway under OOD}
  \label{fig:sway-rate-ood}
\end{subfigure}
\hfill
\begin{subfigure}[b]{0.45\columnwidth}
  \centering
  \includegraphics[width=\textwidth]{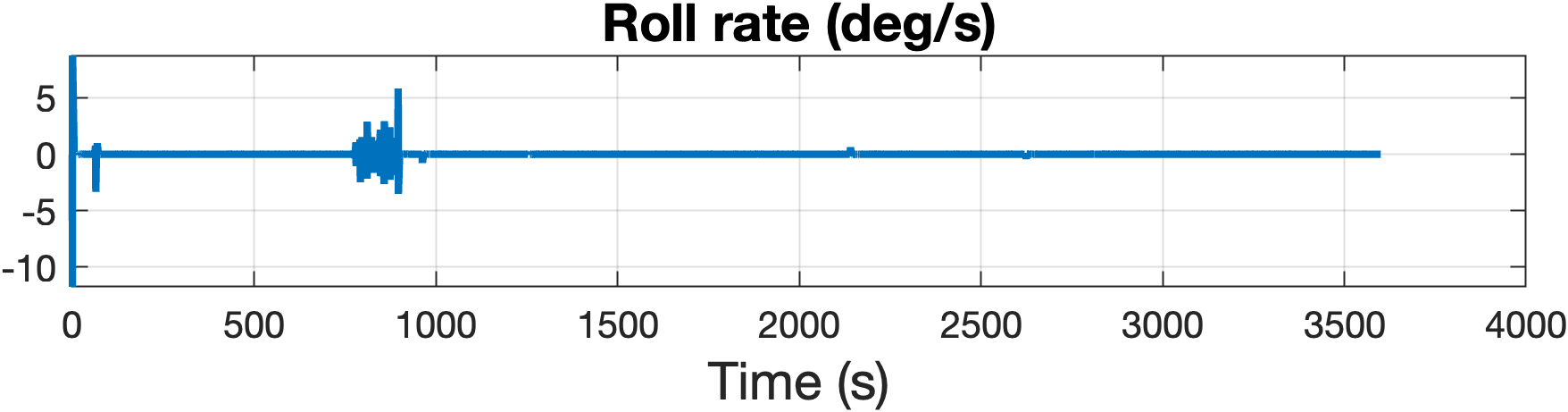}
  \caption{Roll under OOD}
  \label{fig:roll-rate-ood}
\end{subfigure}
\caption{Effects of sway (a) and roll (b) under OOD conditions, i.e., increased ocean current speed at a random time.}
\label{fig:combined-ood-effects}
\end{figure}

The high accuracy of \approach{} in detecting OOD occurrences under environmental disturbances can be attributed to the nature of the data and the distinct impact of such disturbances on the vessel's motions. During simulation, an environmental disturbance, like a sudden spike in ocean current, visibly impacts multiple aspects of the vessel’s dynamics—especially sway, yaw, and roll. 
As shown in \cref{fig:combined-ood-effects}, the disturbance creates sharp deviations in sway velocity (\cref{fig:sway-rate-ood}) and roll rate (\cref{fig:roll-rate-ood}), producing distinct patterns that stand out from normal behavior. These coordinated changes make it easier for \approach{} to detect OOD events, as the disturbance affects multiple motion parameters simultaneously, creating a clear signature of abnormality.

\begin{myexamplec}{\textit{\textbf{RQ3 Summary:}}}
\approach{} effectively detects OOD events caused by environmental disturbances like ocean currents. Sharp, simultaneous deviations in sway, yaw, and roll create clear patterns that make these disturbances easy to distinguish from normal behavior, resulting in near-perfect detection accuracy across all vessels tested.
\end{myexamplec}

\subsubsection{RQ4 Results (Statistical Comparison)}






\paragraph{\textbf{a) Sensor Noise.}}
\cref{tab:rq3_sensornoise} presents the results of the comparative analysis among OOD detection approaches based on sensor noise. 
It can be observed that \emph{p-values} obtained from the Chi-square test are lower than 0.001 for all comparisons across all vessels. 
This indicates statistically significant differences among all approaches. 
The results from Cohen's effect size analysis reveal that \approach{} consistently outperforms DTM-R across all vessels. 
When comparing \approach{} to DTM-E, \approach{} outperforms for three vessels, namely Mariner, NPS AUV, and Otter.
However, in the case of Remus 100, DTM-E performs better than \approach{} with a small margin. 
Moreover, when DTM-R and DTM-E are compared, DTM-R only outperforms for the Mariner vessel, while it underperforms for the other three vessels, i.e., Remus 100, NPS AUV, and Otter.

\begin{table}[h!]
\ra{1.1}
\centering
\scriptsize
\caption{\textbf{RQ4}: Statistical comparison of approaches for OOD detection based on sensor noise}
\resizebox{0.82\textwidth}{!}{%
\begin{tabular}{l c c c c c l}
\toprule
Vessel & Method 1 & Method 2 & Comparison &  Chi-square \textit{p-value} & Cohen's h & Magnitude \\
\midrule
\multirow{3}{*}{Mariner} & \approach{} & DTM-R & Better & < .001 & 0.27 & Small \\
                         & \approach{} & DTM-E & Better & < .001 & 1.52 & Large \\
                         & DTM-R & DTM-E & Better & < .001 & 0.87 & Large \\
\midrule

\multirow{3}{*}{Remus 100} & \approach{} & DTM-R & Better & < .001 & 1.83 & Large \\
                         & \approach{} & DTM-E & Worse & < .001 & 0.20 & Small \\
                         & DTM-R & DTM-E & Worse & < .001 & 2.04 & Large \\

\midrule

\multirow{3}{*}{NPS AUV} & \approach{} & DTM-R & Better & <.001 & 1.61 & Large \\
                         & \approach{} & DTM-E & Better & < .001 & 0.98 & Large \\
                         & DTM-R & DTM-E & Worse & < .001 & 0.63 & Medium \\

\midrule

\multirow{3}{*}{Otter} & \approach{} & DTM-R & Better & < .001 & 2.25 & Large \\
                         & \approach{} & DTM-E & Better & < .001 & 0.15 & Negligible \\
                         & DTM-R & DTM-E & Worse & < .001 & 2.10 & Large \\
                         
\bottomrule
\label{tab:rq3_sensornoise}
\end{tabular}}
\end{table}

\begin{table}[h!]
\ra{1.1}
\centering
\scriptsize
\caption{\textbf{RQ4}: Statistical comparison of approaches for OOD detection based on actuator noise}
\resizebox{0.82\textwidth}{!}{%

\begin{tabular}{l c c c c c l}
\toprule
Vessel & Method 1 & Method 2 & Comparison &  Chi-square \textit{p-value} & Cohen's h & Magnitude \\
\midrule
\multirow{3}{*}{Mariner} & \approach{} & DTM-R & Better & < .001 & 0.20 & Small \\
                         & \approach{} & DTM-E & Better & < .001 & 1.40 & Large \\
                         & DTM-R & DTM-E & Better & < .001 & 1.20 & Large \\
\midrule

\multirow{3}{*}{Container} & \approach{} & DTM-R & Better & < .001 & 0.66 & Medium \\
                         & \approach{} & DTM-E & Better & < .001 & 1.54 & Large \\
                         & DTM-R & DTM-E & Worse & < .001 & 0.87 & Large \\

\midrule

\multirow{3}{*}{Remus 100} & \approach{} & DTM-R & Worse & < .001 & 0.75 & Medium \\
                         & \approach{} & DTM-E & Better & < .001 & 0.60 & Medium \\
                         & DTM-R & DTM-E & Better & < .001 & 1.35 & Medium \\
                         
\bottomrule
\label{tab:rq3_actuatornoise}
\end{tabular}}
\end{table}

\begin{table}[h!]
\ra{1.1}
\centering
\scriptsize
\caption{\textbf{RQ4}: Statistical comparison of approaches for OOD detection based on ocean current}
\resizebox{0.82\textwidth}{!}{%

\begin{tabular}{l c c c c c c}
\toprule
Vessel & Method 1 & Method 2 & Comparison & Chi-square \textit{p-value} & Cohen's h & Magnitude \\
\midrule
\multirow{3}{*}{Remus 100} & \approach{} & DTM-R & Equal & > .05 & 0.11 & Negligible \\
                         & \approach{} & DTM-E & Better & < .001 & 0.52 & Medium \\
                         & DTM-R & DTM-E & Better & < .001 & 0.64 & Medium \\
\midrule

\multirow{3}{*}{NPS AUV} & \approach{} & DTM-R & Equal & > .05 & 0.11 & Negligible \\
                         & \approach{} & DTM-E & Better & < .001 & 0.52 & Medium \\
                         & DTM-R & DTM-E & Worse & < .001 & 0.64 & Medium \\

\midrule

\multirow{3}{*}{Otter} & \approach{} & DTM-R & Equal & > .05 & 0.11 & Negligible \\
                         & \approach{} & DTM-E & Better & < .001 & 0.35 & Small \\
                         & DTM-R & DTM-E & Better & < .001 & 0.46 & Small \\
                         
\bottomrule
\label{tab:rq3_oceancurrent}
\end{tabular}}
\end{table}

\paragraph{\textbf{b) Actuator Noise.}}
For the results related to actuator noise, \cref{tab:rq3_actuatornoise} presents the comparison among all approaches.
The results indicate that the \emph{p-values} derived from the Chi-square test for all comparisons across each vessel are below 0.001, suggesting statistically significant differences among all the approaches. 
Based on Cohen's effect size comparison, \approach{} outperforms both DTM-R and DTM-E for the Mariner and Container vessels. 
However, for the Remus 100 vessel, while \approach{} outperforms DTM-E, it underperforms in comparison to DTM-R, though with a medium magnitude.
When comparing DTM-R and DTM-E, DTM-R performs better than DTM-E in the case of Mariner and Remus 100, whereas it underperforms for the Container vessel. 

\paragraph{\textbf{c) Environmental Disturbances.}}
When comparing approaches based on environmental disturbances, specifically ocean current, \cref{tab:rq3_oceancurrent} presents the results for the Remus 100, NPS AUV, and Otter vessels. 
In the comparison between \approach{} and DTM-R, the Chi-square \emph{p-values} exceed 0.05 for all three vessels, demonstrating that no statistically significant differences exist between them. 
Furthermore, Cohen's effect size suggests a \emph{negligible} magnitude of improvement. 
As a result, we consider the performance of both \approach{} and DTM-R nearly equivalent.
For the comparison between \approach{} and DTM-E, the Chi-square \emph{p-values} are lower than 0.001 in all cases, indicating statistically significant differences. 
The results for Cohen's effect size suggest that \approach{} performs better than DTM-E for all vessels. 
When comparing DTM-R and DTM-E, DTM-R outperforms DTM-E in all cases.


\begin{myexamplec}{\textit{\textbf{RQ4 Summary:}}}   
\approach{} consistently outperformed DTM-R and DTM-E in OOD detection across sensor noise, actuator noise, and environmental disturbances for most vessels. The only exceptions were with Remus 100, where \approach{} underperformed under sensor and actuator noise. Between DTM-R and DTM-E, DTM-R performed better with actuator noise and environmental disturbances, while DTM-E outperformed under sensor noise.
\end{myexamplec}

\subsection{Threats to Validity}
To address threats to \textit{external validity}, we evaluated \approach{} using five different AVs with varying characteristics, such as various DoF. 
Furthermore, these AVs represent a range of vessel sizes---small (e.g., Remus 100 and Otter), medium (e.g., NPS AUV), and large (e.g., Container and Mariner)---ensuring a diverse and representative sample.
Although our experimental results may not generalize to all types of AVs, this limitation is a common concern in empirical research~\cite{lago2024threats}. 

The potential threats to \textit{internal validity} may occur due to the selection of model architectures, hyperparameters, ReLiNet usage, and simulator configurations. 
To mitigate these threats, we conducted a pilot study to search for optimal model architecture and hyperparameters.  
For ReLiNet, we used the latest release available from the online repository and integrated it into our approach with its default settings. 
Additionally, we adhered to the default simulator parameters while defining AV maneuvers and running simulations.
Another internal validity concern is \approach{}’s reliance on historical distribution data for detecting OOD events. This dependence may limit the system’s ability to adapt to novel situations or unexpected environmental conditions not captured in the training data. While including diverse AV types and environmental conditions mitigates this limitation, the approach may struggle in scenarios diverging significantly from the training distribution. Future work will explore adaptive methods, such as fine-tuning or online learning, to enhance robustness against unseen conditions. 

To reduce the \textit{construct validity threats}, we carefully selected evaluation metrics and statistical tests adhering to recommended practices. 
For RQ1 and RQ2 results, we used AUROC and TNR@TPR95, both commonly applied in OOD detection studies~\cite{che2021deep, hsu2020generalized}. 
For RQ3, we employed the Chi-square test and Cohen’s h effect size measure, following the well-established guidelines~\cite{arcuri2011practical}. 
A potential threat to {conclusion validity} is the randomness in model predictions. To address this, we followed standard practices, using a 70-30 training-testing split in our simulation scenarios. While uncertainty is inherent in ML models, further quantifying it in RNN and Autoencoder models is a promising direction for future work.

\section{Discussion}\label{sec:discussion}
\approach{} demonstrates its effectiveness in achieving high accuracy in OOD detection across different vessels and maneuvers under various noise levels and environmental disturbances. \approach{} consistently and robustly identifies OOD states, making it a reliable tool for enhancing autonomous vessel operations. A key factor in \approach{}'s success is its adaptability to vessel-specific dynamics. Each vessel presents unique characteristics---such as maneuverability and motion patterns of smaller vessels like Remus 100---and \approach{} effectively accommodates these variations. By enabling real-time detection of potentially anomalous states, \approach{} provides operators with timely alerts that allow proactive responses to emerging deviations, reducing the risk of failures. In testing and development, \approach{} proves equally beneficial, supporting scenario-centered testing by enabling systematic simulation, monitoring, and evaluation of high-risk OOD events. Allowing researchers and developers to create controlled, realistic environments where specific OOD conditions, e.g., sensor malfunctions, can be introduced to assess the system's robustness.
Additionally, \approach{} can serve as an input to assess and enhance path-following guidance algorithms, providing real-time alerts that signal the need for course corrections before OOD events occur. 

While the correlation between higher corruption (noise) levels and increased OOD detection is well-known, our findings demonstrate that the proposed approach performs consistently well across different levels of sensor noise, though its effectiveness varies across vessels. For example, by adding Gaussian noise to the \textit{x} and \textit{y} positions in 2D navigation scenarios, we observed robust OOD detection performance with minor variability in the results. This highlights the effectiveness of our method in handling varying levels of corruption.

However, our correlation analysis reveals that while moderate noise contributes positively to OOD detection, higher levels of corruption do not always enhance performance. In vessels such as Remus 100, we observed a strong negative correlation between extreme rudder angles and OOD detection metrics, indicating that excessive actuator noise may introduce unstructured motion variability that reduces \approach{}’s ability to distinguish OOD states from normal operations.
In contrast, when analyzing the impact of environmental disturbances such as ocean currents, we found that \approach{} maintains high detection accuracy across all tested vessels, as reflected in near-perfect AUROC and TNR@TPR95 scores. Unlike actuator noise, which can create unpredictable deviations, ocean currents affect multiple motion parameters in a more structured manner, producing clear behavioral shifts that the model can effectively learn to recognize.
Overall, while \approach{} demonstrates strong performance, some limitations remain. 
Its effectiveness is vessel-dependent, as different AVs exhibit varying sensitivities to sensor and actuator noise, influencing detection accuracy.
Additionally, \approach{}'s effectiveness is highly dependent on the quality and coverage of training data, which may impact detection in untrained scenarios. Adapting \approach{} to diverse AV types may require customization to account for vessel-specific dynamics, and detecting entirely novel OOD events could necessitate ongoing retraining. 





\section{Conclusions and Future Work}\label{sec:conclusion}
Autonomous Vessels (AVs) are complex cyber-physical systems with substantial software implemented for various functionality such as path planning. During their operation, it is important to ensure that when AV follows a path to its destination, its state does not go out-of-distribution (OOD) since it can represent potentially unsafe behavior. To this end, we present a digital twin-based approach to detect such OOD in real-time before it happens to enable a shipmaster to take necessary actions if needed. The approach is data-driven, i.e., we built digital twins as machine learning models that can predict the future state of an AV together with whether the future state could potentially be OOD. We experimented with five vessels across waypoint and zigzag maneuvering under simulated conditions, including sensor and actuator noise, and environmental disturbances, i.e., ocean current. \approach{} demonstrated high accuracy in detecting OOD states, achieving AUROC and TNR@TPR95 scores up to 99\% across multiple vessels. 
Furthermore, our comparison with DTM-R and DTM-E methods showed that \approach{} consistently outperformed both approaches in OOD detection across most scenarios and vessels. However, some performance variations across vessels and noise conditions suggest that OOD detection effectiveness is influenced by vessel-specific characteristics. These findings highlight the need for further research into adaptive, vessel-specific tuning strategies to enhance detection robustness across diverse maritime environments.
In the future, we plan to include industrial AV case studies with real operational data from AVs. 
We will also focus on optimizing the current models and systematically evaluating alternative approaches (e.g., transformer neural networks). 
Finally, we aim to extend our approach for uncertainty quantification in OOD detection to further improve digital twins' capability.

\textbf{Replication package:} For replicability, we provide a data package, including code, datasets, and analysis scripts at Zenodo: https://doi.org/10.5281/zenodo.14019147. 


\section*{Acknowledgments}

The Norwegian Ministry of Education and Research supports Erblin Isaku's PhD work reported in this paper. 
The work is also supported by the RoboSAPIENS project funded by the European Commission’s Horizon Europe programme under grant agreement number 101133807.

\bibliography{references}  

\newpage
\appendix
\section{Appendix: Additional Results}

\begin{table*}[h]
\centering
\caption{\textbf{RQ1a:} Statistical comparison of \approach{} across vessel pairs using AUROC and TNR@TPR95. We assess whether distributions differ significantly using rank-based tests, followed by Dunn’s test for pairwise comparisons. Results include the p-value, Vargha-Delaney (\vda{}) effect size, and the comparative outcome (Comp.). A result is statistically significant if \textit{p $<$ 0.05} (marked with \textsuperscript{\(\star\)}) and the effect size is non-negligible (mark different from \textsuperscript{\(\bullet\)}). Effect sizes are categorized as Negligible (\(\bullet\)), Small (\(\triangledown\)), Medium (\(\triangle\)), and Large (\(\diamond\)). The Comp. column indicates whether \approach{} performed better, worse, or showed no significant difference between 1\textsuperscript{st} vessel and 2\textsuperscript{nd} vessel model.}
\label{tab:rq1a_oddit_statistical_results}
\scalebox{1.1}{
\begin{tabular}{llccl ccl}
\toprule
\multirow{2}{*}{Vessel 1}  & \multirow{2}{*}{Vessel 2}  & \multicolumn{3}{c}{AUROC} & \multicolumn{3}{c}{TNR@TPR95} \\
\cmidrule(lr){3-5} \cmidrule(lr){6-8}
& & p-value & \vda{} Effect & Comp. & p-value & \vda{} Effect & Comp. \\
\midrule
Mariner  & Remus 100  & $<$ 0.05\textsuperscript{\(\star\)}  & 0.057  & Worse\textsuperscript{\(\diamond\)}  & $<$ 0.05\textsuperscript{\(\star\)}  & 0.043  & Worse\textsuperscript{\(\diamond\)}  \\
Mariner  & NPS AUV  & $<$ 0.05\textsuperscript{\(\star\)}  & 0.107  & Worse\textsuperscript{\(\diamond\)}  & $<$ 0.05\textsuperscript{\(\star\)}  & 0.087  & Worse\textsuperscript{\(\diamond\)}  \\
Mariner  & Otter  & $<$ 0.05\textsuperscript{\(\star\)}  & 0.086  & Worse\textsuperscript{\(\diamond\)}  & $<$ 0.05\textsuperscript{\(\star\)}  & 0.064 & Worse\textsuperscript{\(\diamond\)}   \\
Remus 100  & NPS AUV  & $<$ 0.05\textsuperscript{\(\star\)}  & 0.726  & Better\textsuperscript{\(\triangle\)}  & $<$ 0.05\textsuperscript{\(\star\)}  & 0.778  & Better\textsuperscript{\(\diamond\)}  \\
Remus 100  & Otter  & $<$ 0.05\textsuperscript{\(\star\)}  & 0.740  & Better\textsuperscript{\(\diamond\)}  & $<$ 0.05\textsuperscript{\(\star\)}  & 0.810  & Better\textsuperscript{\(\diamond\)}  \\
NPS AUV  & Otter  & $>$ 0.05  & 0.503  & Equal\textsuperscript{\(\bullet\)}  & $>$ 0.05  & 0.527  & Equal\textsuperscript{\(\bullet\)}  \\
\hline
\end{tabular}
}
\end{table*}


\begin{table}[h]
    \centering
    \caption{\textbf{RQ1b:} Statistical comparison of noise magnitudes for NPS AUV vessel using AUROC and TNR@TPR95. We assess whether distributions differ significantly using rank-based tests, followed by Dunn’s test for pairwise comparisons. Results include the p-value, Vargha-Delaney (\vda{}) effect size, and the comparative outcome (Comp.). A result is statistically significant if \textit{p $<$ 0.05} (marked with \textsuperscript{\(\star\)}) and the effect size is non-negligible (mark different from \textsuperscript{\(\bullet\)}). Effect sizes are categorized as Negligible (\(\bullet\)), Small (\(\triangledown\)), Medium (\(\triangle\)), and Large (\(\diamond\)).}
    \label{tab:rq1b_noise_magnitude_comparison_npsauv}
    \begin{tabular}{ll ccl ccl}
        \toprule
        \multirow{2}{*}{Noise 1} & \multirow{2}{*}{Noise 2} & \multicolumn{3}{c}{AUROC} & \multicolumn{3}{c}{TNR@TPR95} \\ 
        \cmidrule(lr){3-5} \cmidrule(lr){6-8} 
        & & p-value & \vda{} Effect & Comp. & p-value & \vda{} Effect & Comp. \\ 
        \midrule
         m2 & m3 & > 0.05 & 0.24 & Worse\textsuperscript{\(\diamond\)} & > 0.05 & 0.23 & Worse\textsuperscript{\(\diamond\)} \\
         m2 & m4 & > 0.05 & 0.12 & Worse\textsuperscript{\(\diamond\)} & > 0.05 & 0.09 & Worse\textsuperscript{\(\diamond\)} \\
         m2 & m5 & > 0.05 & 0.06 & Worse\textsuperscript{\(\diamond\)} & < 0.05\textsuperscript{\(\star\)} & 0.05 & Worse\textsuperscript{\(\diamond\)} \\
         m2 & m6 & < 0.05\textsuperscript{\(\star\)} & 0.05 & Worse\textsuperscript{\(\diamond\)} & < 0.05\textsuperscript{\(\star\)} & 0.02 & Worse\textsuperscript{\(\diamond\)} \\
         m2 & m7 & < 0.05\textsuperscript{\(\star\)} & 0.02 & Worse\textsuperscript{\(\diamond\)} & < 0.05\textsuperscript{\(\star\)} & 0.00 & Worse\textsuperscript{\(\diamond\)} \\
         m2 & m8 & < 0.05\textsuperscript{\(\star\)} & 0.01 & Worse\textsuperscript{\(\diamond\)} & < 0.05\textsuperscript{\(\star\)} & 0.00 & Worse\textsuperscript{\(\diamond\)} \\
         m3 & m4 & > 0.05 & 0.34 & Worse\textsuperscript{\(\triangledown\)} & > 0.05 & 0.26 & Worse\textsuperscript{\(\diamond\)} \\
         m3 & m5 & > 0.05 & 0.26 & Worse\textsuperscript{\(\diamond\)} & > 0.05 & 0.16 & Worse\textsuperscript{\(\diamond\)} \\
         m3 & m6 & > 0.05 & 0.18 & Worse\textsuperscript{\(\diamond\)} & < 0.05\textsuperscript{\(\star\)} & 0.08 & Worse\textsuperscript{\(\diamond\)} \\
         m3 & m7 & > 0.05 & 0.12 & Worse\textsuperscript{\(\diamond\)} & > 0.05 & 0.12 & Worse\textsuperscript{\(\diamond\)} \\
         m3 & m8 & < 0.05\textsuperscript{\(\star\)} & 0.07 & Worse\textsuperscript{\(\diamond\)} & < 0.05\textsuperscript{\(\star\)} & 0.04 & Worse\textsuperscript{\(\diamond\)} \\
         m4 & m5 & > 0.05 & 0.41 & Worse\textsuperscript{\(\triangledown\)} & > 0.05 & 0.42 & Worse\textsuperscript{\(\triangledown\)} \\
         m4 & m6 & > 0.05 & 0.30 & Worse\textsuperscript{\(\triangle\)} & > 0.05 & 0.28 & Worse\textsuperscript{\(\triangle\)} \\
         m4 & m7 & > 0.05 & 0.21 & Worse\textsuperscript{\(\diamond\)} & > 0.05 & 0.27 & Worse\textsuperscript{\(\triangle\)} \\
         m4 & m8 & > 0.05 & 0.18 & Worse\textsuperscript{\(\diamond\)} & > 0.05 & 0.22 & Worse\textsuperscript{\(\diamond\)} \\
         m5 & m6 & > 0.05 & 0.37 & Worse\textsuperscript{\(\triangledown\)} & > 0.05 & 0.32 & Worse\textsuperscript{\(\triangle\)} \\
         m5 & m7 & > 0.05 & 0.31 & Worse\textsuperscript{\(\triangle\)} & > 0.05 & 0.38 & Worse\textsuperscript{\(\triangledown\)} \\
         m5 & m8 & > 0.05 & 0.19 & Worse\textsuperscript{\(\diamond\)} & > 0.05 & 0.20 & Worse\textsuperscript{\(\diamond\)} \\
         m6 & m7 & > 0.05 & 0.41 & Worse\textsuperscript{\(\triangledown\)} & > 0.05 & 0.54 & Better\textsuperscript{\(\bullet\)} \\
         m6 & m8 & > 0.05 & 0.32 & Worse\textsuperscript{\(\triangle\)} & > 0.05 & 0.44 & Worse\textsuperscript{\(\bullet\)} \\
         m7 & m8 & > 0.05 & 0.43 & Worse\textsuperscript{\(\bullet\)} & > 0.05 & 0.34 & Worse\textsuperscript{\(\triangledown\)} \\
        \bottomrule
    \end{tabular}
\end{table}

\begin{table}[h]
    \centering
    \caption{\textbf{RQ1b:} Statistical comparison of noise magnitudes for Otter vessel using AUROC and TNR@TPR95. We assess whether distributions differ significantly using rank-based tests, followed by Dunn’s test for pairwise comparisons. Results include the p-value, Vargha-Delaney (\vda{}) effect size, and the comparative outcome (Comp.). A result is statistically significant if \textit{p $<$ 0.05} (marked with \textsuperscript{\(\star\)}) and the effect size is non-negligible (mark different from \textsuperscript{\(\bullet\)}). Effect sizes are categorized as Negligible (\(\bullet\)), Small (\(\triangledown\)), Medium (\(\triangle\)), and Large (\(\diamond\)).}
    \label{tab:rq1b_noise_magnitude_comparison_otter}
    \begin{tabular}{ll ccl ccl}
        \toprule
        \multirow{2}{*}{Noise 1} & \multirow{2}{*}{Noise 2} & \multicolumn{3}{c}{AUROC} & \multicolumn{3}{c}{TNR@TPR95} \\ 
        \cmidrule(lr){3-5} \cmidrule(lr){6-8} 
        & & p-value & \vda{} Effect & Comp. & p-value & \vda{} Effect & Comp. \\ 
        \midrule
         m2 & m3 & > 0.05 & 0.29 & Worse\textsuperscript{\(\triangle\)} & > 0.05 & 0.32 & Worse\textsuperscript{\(\triangle\)} \\
         m2 & m4 & > 0.05 & 0.14 & Worse\textsuperscript{\(\diamond\)} & > 0.05 & 0.20 & Worse\textsuperscript{\(\diamond\)} \\
         m2 & m5 & > 0.05 & 0.13 & Worse\textsuperscript{\(\diamond\)} & > 0.05 & 0.20 & Worse\textsuperscript{\(\diamond\)} \\
         m2 & m6 & > 0.05 & 0.10 & Worse\textsuperscript{\(\diamond\)} & > 0.05 & 0.11 & Worse\textsuperscript{\(\diamond\)} \\
         m2 & m7 & < 0.05\textsuperscript{\(\star\)} & 0.06 & Worse\textsuperscript{\(\diamond\)} & < 0.05\textsuperscript{\(\star\)} & 0.07 & Worse\textsuperscript{\(\diamond\)} \\
         m2 & m8 & < 0.05\textsuperscript{\(\star\)} & 0.02 & Worse\textsuperscript{\(\diamond\)} & < 0.05\textsuperscript{\(\star\)} & 0.06 & Worse\textsuperscript{\(\diamond\)} \\
         m3 & m4 & > 0.05 & 0.29 & Worse\textsuperscript{\(\triangle\)} & > 0.05 & 0.36 & Worse\textsuperscript{\(\triangledown\)} \\
         m3 & m5 & > 0.05 & 0.28 & Worse\textsuperscript{\(\triangle\)} & > 0.05 & 0.33 & Worse\textsuperscript{\(\triangle\)} \\
         m3 & m6 & > 0.05 & 0.23 & Worse\textsuperscript{\(\diamond\)} & > 0.05 & 0.23 & Worse\textsuperscript{\(\diamond\)} \\
         m3 & m7 & > 0.05 & 0.15 & Worse\textsuperscript{\(\diamond\)} & > 0.05 & 0.15 & Worse\textsuperscript{\(\diamond\)} \\
         m3 & m8 & < 0.05\textsuperscript{\(\star\)} & 0.07 & Worse\textsuperscript{\(\diamond\)} & < 0.05\textsuperscript{\(\star\)} & 0.12 & Worse\textsuperscript{\(\diamond\)} \\
         m4 & m5 & > 0.05 & 0.39 & Worse\textsuperscript{\(\triangledown\)} & > 0.05 & 0.48 & Worse\textsuperscript{\(\bullet\)} \\
         m4 & m6 & > 0.05 & 0.38 & Worse\textsuperscript{\(\triangledown\)} & > 0.05 & 0.36 & Worse\textsuperscript{\(\triangledown\)} \\
         m4 & m7 & > 0.05 & 0.31 & Worse\textsuperscript{\(\triangle\)} & > 0.05 & 0.26 & Worse\textsuperscript{\(\diamond\)} \\
         m4 & m8 & > 0.05 & 0.19 & Worse\textsuperscript{\(\diamond\)} & > 0.05 & 0.23 & Worse\textsuperscript{\(\diamond\)} \\
         m5 & m6 & > 0.05 & 0.51 & Better\textsuperscript{\(\bullet\)} & > 0.05 & 0.39 & Worse\textsuperscript{\(\triangledown\)} \\
         m5 & m7 & > 0.05 & 0.40 & Worse\textsuperscript{\(\triangledown\)} & > 0.05 & 0.27 & Worse\textsuperscript{\(\triangle\)} \\
         m5 & m8 & > 0.05 & 0.32 & Worse\textsuperscript{\(\triangle\)} & > 0.05 & 0.26 & Worse\textsuperscript{\(\diamond\)} \\
         m6 & m7 & > 0.05 & 0.41 & Worse\textsuperscript{\(\triangledown\)} & > 0.05 & 0.35 & Worse\textsuperscript{\(\triangledown\)} \\
         m6 & m8 & > 0.05 & 0.29 & Worse\textsuperscript{\(\triangle\)} & > 0.05 & 0.21 & Worse\textsuperscript{\(\diamond\)} \\
         m7 & m8 & > 0.05 & 0.42 & Worse\textsuperscript{\(\triangledown\)} & > 0.05 & 0.35 & Worse\textsuperscript{\(\triangledown\)} \\
        \bottomrule
    \end{tabular}
\end{table}

\begin{table}
\scriptsize
\ra{1.1}
\caption{
\textbf{RQ1c:} Statistical comparison of \approach{} across vessel pairs and noise magnitudes using AUROC and TNR@TPR95. We assess whether distributions differ significantly using rank-based tests, followed by Dunn’s test for pairwise comparisons. Results include the p-value, Vargha-Delaney (\vda{}) effect size, and the comparative outcome (Comp.). A result is statistically significant if \textit{p $<$ 0.05} (marked with \textsuperscript{\(\star\)}) and the effect size is non-negligible (mark different from \textsuperscript{\(\bullet\)}). Effect sizes are categorized as Negligible (\(\bullet\)), Small (\(\triangledown\)), Medium (\(\triangle\)), and Large (\(\diamond\)). The Comp. column indicates whether \approach{} performed better, worse, or showed no significant difference between 1\textsuperscript{st} vessel and 2\textsuperscript{nd} vessel model.
}
\label{tab:rq1c_statistical_results}
\centering
\scalebox{1.2}{
\begin{tabular}{l ll ccl ccl}
\toprule
\multirow{3}{*}{Noise Mag.} & \multirow{3}{*}{Vessel 1} & \multirow{3}{*}{Vessel 2} & \multicolumn{3}{c}{AUROC} & \multicolumn{3}{c}{TNR@TPR95} \\ 
\cmidrule(lr){4-6} \cmidrule(lr){7-9}
 & & & p-value & \vda{} Effect & Comp. & p-value & \vda{} Effect & Comp. \\ 
\midrule
\multirow{6}{*}{m2} 
 & Mariner & Remus 100  & $<$ 0.05\textsuperscript{\(\star\)} & 0.18 & Worse\textsuperscript{\(\diamond\)} & $<$ 0.05\textsuperscript{\(\star\)} & 0.15 & Worse\textsuperscript{\(\diamond\)} \\
 & Mariner & NPS AUV & $>$ 0.05 & 0.36 & Worse\textsuperscript{\(\triangledown\)} & $>$ 0.05 & 0.30 & Worse\textsuperscript{\(\triangledown\)} \\
 & Mariner & Otter  & $>$ 0.05 & 0.25 & Worse\textsuperscript{\(\diamond\)} & $>$ 0.05 & 0.18 & Worse\textsuperscript{\(\diamond\)} \\
 & Remus 100   & NPS AUV & $>$ 0.05 & 0.83 & Better\textsuperscript{\(\diamond\)} & $>$ 0.05 & 0.79 & Better\textsuperscript{\(\diamond\)} \\
 & Remus 100   & Otter  & $>$ 0.05 & 0.76 & Better\textsuperscript{\(\diamond\)} & $>$ 0.05 & 0.69 & Better\textsuperscript{\(\triangledown\)} \\
 & NPS AUV  & Otter  & $>$ 0.05 & 0.35 & Worse\textsuperscript{\(\triangledown\)} & $>$ 0.05 & 0.28 & Worse\textsuperscript{\(\triangledown\)} \\
\midrule
\multirow{6}{*}{m3} 
 & Mariner & Remus 100  & $<$ 0.05\textsuperscript{\(\star\)} & 0.03 & Worse\textsuperscript{\(\diamond\)} & $<$ 0.05\textsuperscript{\(\star\)} & 0.04 & Worse\textsuperscript{\(\diamond\)} \\
 & Mariner & NPS AUV & $>$ 0.05 & 0.09 & Worse\textsuperscript{\(\diamond\)} & $>$ 0.05 & 0.14 & Worse\textsuperscript{\(\diamond\)} \\
 & Mariner & Otter  & $<$ 0.05\textsuperscript{\(\star\)} & 0.09 & Worse\textsuperscript{\(\diamond\)} & $<$ 0.05\textsuperscript{\(\star\)} & 0.04 & Worse\textsuperscript{\(\diamond\)} \\
 & Remus 100   & NPS AUV & $>$ 0.05 & 0.84 & Better\textsuperscript{\(\diamond\)} & $>$ 0.05 & 0.84 & Better\textsuperscript{\(\diamond\)} \\
 & Remus 100   & Otter  & $>$ 0.05 & 0.81 & Better\textsuperscript{\(\diamond\)} & $>$ 0.05 & 0.81 & Better\textsuperscript{\(\diamond\)} \\
 & NPS AUV  & Otter  & $>$ 0.05 & 0.44 & Equal\textsuperscript{\(\bullet\)} & $>$ 0.05 & 0.35 & Worse\textsuperscript{\(\triangledown\)} \\
\midrule
\multirow{6}{*}{m4} 
 & Mariner & Remus 100  & $<$ 0.05\textsuperscript{\(\star\)} & 0.03 & Worse\textsuperscript{\(\diamond\)} & $<$ 0.05\textsuperscript{\(\star\)} & 0.02 & Worse\textsuperscript{\(\diamond\)} \\
 & Mariner & NPS AUV & $<$ 0.05\textsuperscript{\(\star\)} & 0.07 & Worse\textsuperscript{\(\diamond\)} & $<$ 0.05\textsuperscript{\(\star\)} & 0.02 & Worse\textsuperscript{\(\diamond\)} \\
 & Mariner & Otter  & $<$ 0.05\textsuperscript{\(\star\)} & 0.04 & Worse\textsuperscript{\(\diamond\)} & $<$ 0.05\textsuperscript{\(\star\)} & 0.01 & Worse\textsuperscript{\(\diamond\)} \\
 & Remus 100   & NPS AUV & $>$ 0.05 & 0.77 & Better\textsuperscript{\(\diamond\)} & $>$ 0.05 & 0.83 & Better\textsuperscript{\(\diamond\)} \\
 & Remus 100   & Otter  & $>$ 0.05 & 0.74 & Better\textsuperscript{\(\diamond\)} & $>$ 0.05 & 0.88 & Better\textsuperscript{\(\diamond\)} \\
 & NPS AUV  & Otter  & $>$ 0.05 & 0.42 & Worse\textsuperscript{\(\triangledown\)} & $>$ 0.05 & 0.47 & Equal\textsuperscript{\(\bullet\)} \\
\midrule
\multirow{6}{*}{m5} 
 & Mariner & Remus 100  & $<$ 0.05\textsuperscript{\(\star\)} & 0.04 & Worse\textsuperscript{\(\diamond\)} & $<$ 0.05\textsuperscript{\(\star\)} & 0.02 & Worse\textsuperscript{\(\diamond\)} \\
 & Mariner & NPS AUV & $<$ 0.05\textsuperscript{\(\star\)} & 0.09 & Worse\textsuperscript{\(\diamond\)} & $<$ 0.05\textsuperscript{\(\star\)} & 0.11 & Worse\textsuperscript{\(\diamond\)} \\
 & Mariner & Otter  & $<$ 0.05\textsuperscript{\(\star\)} & 0.08 & Worse\textsuperscript{\(\diamond\)} & $>$ 0.05 & 0.13 & Worse\textsuperscript{\(\diamond\)} \\
 & Remus 100   & NPS AUV & $>$ 0.05 & 0.69 & Better\textsuperscript{\(\triangle\)} & $>$ 0.05 & 0.80 & Better\textsuperscript{\(\diamond\)} \\
 & Remus 100   & Otter  & $>$ 0.05 & 0.73 & Better\textsuperscript{\(\triangle\)} & $>$ 0.05 & 0.83 & Better\textsuperscript{\(\diamond\)} \\
 & NPS AUV  & Otter  & $>$ 0.05 & 0.44 & Equal\textsuperscript{\(\bullet\)} & $>$ 0.05 & 0.61 & Better\textsuperscript{\(\triangledown\)} \\
\midrule
\multirow{6}{*}{m6} 
 & Mariner & Remus 100  & $<$ 0.05\textsuperscript{\(\star\)} & 0.06 & Worse\textsuperscript{\(\diamond\)} & $<$ 0.05\textsuperscript{\(\star\)} & 0.06 & Worse\textsuperscript{\(\diamond\)} \\
 & Mariner & NPS AUV & $>$ 0.05 & 0.15 & Worse\textsuperscript{\(\diamond\)} & $<$ 0.05\textsuperscript{\(\star\)} & 0.08 & Worse\textsuperscript{\(\diamond\)} \\
 & Mariner & Otter  & $>$ 0.05 & 0.14 & Worse\textsuperscript{\(\diamond\)} & $>$ 0.05 & 0.11 & Worse\textsuperscript{\(\diamond\)} \\
 & Remus 100   & NPS AUV & $>$ 0.05 & 0.76 & Better\textsuperscript{\(\diamond\)} & $>$ 0.05 & 0.77 & Better\textsuperscript{\(\diamond\)} \\
 & Remus 100   & Otter  & $>$ 0.05 & 0.74 & Better\textsuperscript{\(\diamond\)} & $>$ 0.05 & 0.84 & Better\textsuperscript{\(\diamond\)} \\
 & NPS AUV  & Otter  & $>$ 0.05 & 0.60 & Better\textsuperscript{\(\triangledown\)} & $>$ 0.05 & 0.76 & Better\textsuperscript{\(\diamond\)} \\
\midrule
\multirow{6}{*}{m7} 
 & Mariner & Remus 100  & $<$ 0.05\textsuperscript{\(\star\)} & 0.01 & Worse\textsuperscript{\(\diamond\)} & $<$ 0.05\textsuperscript{\(\star\)} & 0.00 & Worse\textsuperscript{\(\diamond\)} \\
 & Mariner & NPS AUV & $<$ 0.05\textsuperscript{\(\star\)} & 0.02 & Worse\textsuperscript{\(\diamond\)} & $<$ 0.05\textsuperscript{\(\star\)} & 0.00 & Worse\textsuperscript{\(\diamond\)} \\
 & Mariner & Otter  & $<$ 0.05\textsuperscript{\(\star\)} & 0.01 & Worse\textsuperscript{\(\diamond\)} & $<$ 0.05\textsuperscript{\(\star\)} & 0.00 & Worse\textsuperscript{\(\diamond\)} \\
 & Remus 100   & NPS AUV & $>$ 0.05 & 0.75 & Better\textsuperscript{\(\diamond\)} & $>$ 0.05 & 0.81 & Better\textsuperscript{\(\diamond\)} \\
 & Remus 100   & Otter  & $>$ 0.05 & 0.81 & Better\textsuperscript{\(\diamond\)} & $>$ 0.05 & 0.83 & Better\textsuperscript{\(\diamond\)} \\
 & NPS AUV  & Otter  & $>$ 0.05 & 0.61 & Better\textsuperscript{\(\triangledown\)} & $>$ 0.05 & 0.53 & Equal\textsuperscript{\(\bullet\)} \\
\midrule
\multirow{6}{*}{m8} 
 & Mariner & Remus 100  & $<$ 0.05\textsuperscript{\(\star\)} & 0.00 & Worse\textsuperscript{\(\diamond\)} & $<$ 0.05\textsuperscript{\(\star\)} & 0.00 & Worse\textsuperscript{\(\diamond\)} \\
 & Mariner & NPS AUV & $<$ 0.05\textsuperscript{\(\star\)} & 0.00 & Worse\textsuperscript{\(\diamond\)} & $<$ 0.05\textsuperscript{\(\star\)} & 0.00 & Worse\textsuperscript{\(\diamond\)} \\
 & Mariner & Otter  & $<$ 0.05\textsuperscript{\(\star\)} & 0.00 & Worse\textsuperscript{\(\diamond\)} & $<$ 0.05\textsuperscript{\(\star\)} & 0.00 & Worse\textsuperscript{\(\diamond\)} \\
 & Remus 100   & NPS AUV & $>$ 0.05 & 0.64 & Better\textsuperscript{\(\triangledown\)} & $>$ 0.05 & 0.77 & Better\textsuperscript{\(\diamond\)} \\
 & Remus 100   & Otter  & $>$ 0.05 & 0.65 & Better\textsuperscript{\(\triangledown\)} & $>$ 0.05 & 0.82 & Better\textsuperscript{\(\diamond\)} \\
 & NPS AUV  & Otter  & $>$ 0.05 & 0.69 & Better\textsuperscript{\(\triangle\)} & $>$ 0.05 & 0.64 & Better\textsuperscript{\(\triangledown\)} \\
\bottomrule
\end{tabular} %
}
\end{table}


\begin{table}[h!]
\centering
\scriptsize
\caption{
\textbf{RQ2c:} Pairwise comparison results of \approach{} across vessel pairs based on rudder angles. Results include the Dunn’s test p-value, Vargha-Delaney (\vda{}) effect size, and the comparative outcome (Comp.). A result is statistically significant if p $<$ 0.05 (marked with \textsuperscript{\(\star\)}) and the effect size is non-negligible (mark different from \textsuperscript{\(\bullet\)}). Effect sizes are categorized as Negligible (\(\bullet\)), Small (\(\triangledown\)), Medium (\(\triangle\)), and Large (\(\diamond\)).}
\label{tab:rq2c_ood_comparison}
\resizebox{\textwidth}{!}{%
\begin{tabular}{l llcccccc}
\toprule
\multirow{2}{*}{Rudder Angle} & \multirow{2}{*}{Vessel 1} & \multirow{2}{*}{Vessel 2} & \multicolumn{3}{c}{AUROC} & \multicolumn{3}{c}{TNR@TPR95} \\ 
\cmidrule(lr){4-6} \cmidrule(lr){7-9}
 & & & p-value & \vda{} Effect & Comp. & p-value & \vda{} Effect & Comp. \\ 
\midrule
\multirow{3}{*}{40°} 
 & Container & Mariner  & $>$ 0.05  & 0.06  & Worse\textsuperscript{\(\diamond\)} & $>$ 0.05  & 0.25  & Worse\textsuperscript{\(\diamond\)} \\
 & Container & Remus 100 & $<$ 0.05\textsuperscript{\(\star\)}  & 0.00  & Worse\textsuperscript{\(\diamond\)} & $<$ 0.05\textsuperscript{\(\star\)}  & 0.00  & Worse\textsuperscript{\(\diamond\)} \\
 & Mariner   & Remus 100 & $>$ 0.05 & 0.37  & Worse\textsuperscript{\(\triangledown\)} & $>$ 0.05  & 0.18  & Worse\textsuperscript{\(\diamond\)} \\
\bottomrule
\end{tabular}%
}
\end{table}

\end{document}